%% file: main.tex
\documentclass[USenglish,oneside,twocolumn]{article}

\input{header}
\graphicspath{{./figures/}}

\DOI{foobar}
\cclogo{\includegraphics{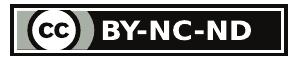}}
\allowdisplaybreaks
\begin{document}
  \author*[1]{Jaewoo Lee}
  \author[2]{Daniel Kifer}
  \affil[1]{University of Georgia, E-mail: jwlee@cs.uga.edu}
  \affil[2]{Penn State University, E-mail: dkifer@cse.psu.edu}

  \title{\huge Scaling up Differentially Private Deep Learning with Fast Per-Example Gradient Clipping}

  \runningtitle{Fast Per-Example Gradient Clipping}
  %\subtitle{...}

  \begin{abstract}{Recent work on Renyi Differential Privacy has shown
      the feasibility of applying differential privacy to deep
      learning tasks. Despite their promise, however, differentially
      private deep networks often lag far behind their non-private
      counterparts in accuracy, showing the need for more research in
      model architectures, optimizers, etc. One of the barriers to
      this expanded research is the training time --- often orders of
      magnitude larger than training non-private networks. The reason
      for this slowdown is a crucial privacy-related step called
      ``per-example gradient clipping'' whose naive implementation
      undoes the benefits of batch training with GPUs. By analyzing
      the back-propagation  equations we derive new methods for
      per-example gradient clipping that are compatible with
      auto-differeniation (e.g., in PyTorch and TensorFlow) and
      provide better GPU utilization. Our implementation in PyTorch
      showed significant training speed-ups (by factors of 54x - 94x for training
      various models with batch sizes of 128). These techniques work for a variety of architectural
      choices including convolutional layers, recurrent networks,
      attention, residual blocks, etc.} 
\end{abstract}
%  \classification[PACS]{}
 % \communicated{...}
 % \dedication{...}

  \journalname{Proceedings on Privacy Enhancing Technologies}
  \DOI{Editor to enter DOI}
  \startpage{1}
  \received{..}
  \revised{..}
  \accepted{..}

  \journalyear{2020}
  \journalvolume{..}
  \journalissue{..}
 
\maketitle
\section{Introduction}
\label{sec:intro}
\input{introduction}

\section{Preliminaries}
\label{sec:preliminaries}
\input{preliminaries}

\section{The Problem with Per-Example Gradient Clipping}
\label{sec:clipping}
\input{clipping}

\section{Related Work}
\label{sec:related_work}
\input{related}

\section{Faster Deep Learning with Differential Privacy}
\label{sec:algorithm}
\input{algorithm}

\section{Experiments}
\label{sec:experiments}
\input{experiments}

\section{Conclusions}
\label{sec:conclusion}
\input{conclusion}

%\clearpage
\bibliographystyle{abbrv}
\bibliography{reference}

\end{document}

%% file: header.tex
\usepackage[utf8]{inputenc}%(only for the pdftex engine)
\usepackage[big]{dgruyter_NEW}
\usepackage{amsmath,amssymb,amsthm}
\usepackage[inline]{enumitem}
\usepackage[table,usenames,dvipsnames]{xcolor}
\usepackage{graphicx}
\usepackage{tikz}
\usetikzlibrary{calc,fit,arrows}
\usetikzlibrary{positioning,intersections}
\usetikzlibrary{backgrounds}
\usetikzlibrary{matrix,chains}
\usetikzlibrary{patterns}
\usepackage{bm}
\usepackage{mathtools}
\usepackage{xspace}
\usepackage{caption}
\usepackage{subcaption}
\usepackage[linesnumbered,ruled,noline]{algorithm2e}
\usepackage{booktabs}
\usepackage{comment}
\usepackage{balance}

\let\en=\ensuremath

\newtheorem{definition}{Definition}

\newtheorem{lemma}{Lemma}

\allowdisplaybreaks

\DeclarePairedDelimiter{\norm}{\lVert}{\rVert}
\DeclarePairedDelimiter{\ip}{\langle}{\rangle}
\DeclarePairedDelimiterX{\divx}[2]{(}{)}{#1\;\delimsize\|\;#2}

\newcommand{\pdv}[2]{\en{\frac{\partial #1}{\partial #2}}}

\DeclareMathOperator{\E}{\mathbb{E}}          % expectation
\DeclareMathOperator{\Divergence}{\mathfrak{D}}

\newcommand{\Div}[0]{\Divergence_{\alpha}\divx}

\DeclareMathOperator*{\argmin}{arg\,min\,}

\DeclareMathOperator{\Range}{range}

\renewcommand{\vec}[1]{\en{\bm{\mathrm{#1}}}}

\newcommand{\mat}[1]{\en{{\bm{\mathrm{#1}}}}}
\newcommand{\grad}[0]{\en{\nabla}}

\newcommand{\R}[0]{\mathbb{R}}

\renewcommand{\th}[0]{\textsuperscript{th}\xspace}

\newcommand{\revision}[1]{\textcolor{black}{#1}}

%% file: introduction.tex
Machine learning models trained on sensitive datasets, such as medical
records, emails, and financial transactions have great value to
society but also pose risks to individuals who contributed their
information to the training data. \revision{Even if the model parameters are not shared,  black-box access to the models can leak private information \cite{membershipinference}}. 
Differential privacy is a promising framework for mitigating such
risks  because of its strong mathematical guarantees and because of 
 recent advances in differentially private training of predictive models. 

Simple models, such as linear regression and logistic regression, which have convenient mathematical structures (e.g., convexity) and relatively few parameters, have been well-studied in the differentially private literature and many privacy preserving training algorithms have been proposed (e.g., \cite{Chaudhuri2011Objpert,Kifer2012erm,Bassily2014PERM,Iyengar2019amp,Zhang2017RR,Wang2017dpsvrg,Lee2018cdp,Chen2019renyi,KNG}). These algorithms are generally fast and accurate compared to their non-private counterparts.

However, state-of-the-art prediction results generally come from deep artificial neural networks with millions of parameters. These models are not convex and hence require different fitting algorithms to ensure privacy \cite{Abadi2016deep,Papernot2018pate,Jordon2019pategan,Yu2019modelpub}. In the non-private case, training is generally accomplished using stochastic gradient descent backed by GPU/TPU hardware accelerators that process multiple training records together in a batch. In the privacy-preserving case, the most generally applicable training algorithm is also a variation of stochastic gradient descent~\cite{Abadi2016deep}. However, its current implementations (e.g.,~\cite{tensorflowprivacy}) are extremely slow because a key step, ``per-example gradient clipping'',\footnote{Essentially, the gradient contribution of each record in a batch must be normalized first (in a nonlinear way), before the contributions are added together. See Section~\ref{sec:clipping} for details.} limits the batch-processing capabilities of GPUs/TPUs, resulting in slowdowns of up to two orders of magnitude.
This slowdown has a direct impact on differentially-private deep learning research, as it becomes expensive even to experiment with differential privacy and different neural network architectures~\cite{shmatfairness}.

In this paper, we show that most of this slowdown can be avoided. By
analyzing how backpropagation computes the gradients, we derive some
tricks for fast per-example-gradient clipping that are easy to
implement and result in speedups of up to 94x over the naive
approach. These methods take advantage of auto-differentiation
features of standard deep learning packages (such as TensorFlow and
PyTorch~\cite{Paszke2017automatic}) and do not require any low-level
programming (our code 
consists of Python wrappers around PyTorch layer objects --- e.g., a
wrapper for fully connected layers, a wrapper for convolutional
layers, etc.). 

We note that Goodfellow \cite{Goodfellow2015efficient} provided a fast per-example gradient clipping method that only applied to fully connected networks. Our results apply to a wider variety of architectures, including convolutional layers, recurrent networks, attention, residual blocks, etc.

In short, our contributions are as follows.
\begin{itemize}[leftmargin=3mm,topsep=0pt]
\item We present methods for efficiently computing
  per-example gradients for different kinds of deep learning models, achieving a speedup of up to 54x to 94x (depending on the model) compared to naive per-example gradient computation on mini-batches of size 128. This allows hardware-accelerated differentially private training of deep learning models to rival the speed of hardware-accelerated non-private training and thus makes differentially private deep learning possible in practical timeframes.
\item The proposed methods \revision{do not require fundamental changes to GPU parallelization. Instead, they} are easy to implement because they take advantage of automatic differentiation capabilities of modern deep learning packages. Our PyTorch wrappers are being prepared for open-source release.
\item As an application of the proposed framework, we demonstrate how
  to train (under R\'enyi differential privacy) a {\scshape Transformer} encoder block~\cite{Vaswani2017attention}, a key component in an architecture that has lead to recent advances in natural language processing.
\item We perform extensive experiments and empirically show the
  effectiveness of approach for differentially private training of
  various kinds of deep neural network models. 
\end{itemize}

The rest of this paper is organized as follows. In Section~\ref{sec:preliminaries}, we define notations and provide
background on differential privacy. Building on these concepts, we describe the per-example gradient clipping problem in Section \ref{sec:clipping}.
We then discuss related work in Section~\ref{sec:related_work}. We present our proposed methods
in Section~\ref{sec:algorithm}, experimental results in
Section~\ref{sec:experiments} and conclusions in Section~\ref{sec:conclusion}.

%%% Local Variables:
%%% mode: latex
%%% TeX-master: "main"
%%% End:

%% file: preliminaries.tex
In this paper, we use upper-letters (e.g., $W$) to represent matrices, bold-face lower-case (e.g., $\mathbf{x}$) to represent vectors and non-bold lower-case (e.g., $y$) to represent scalars. One exception is that $D$ represents a dataset. Tensors of order 3 or higher (i.e., multidimensional arrays that are indexed by 3 or more variables) are represented in calligraphic font (e.g,. $\mathcal{W}$).

We index vectors using square brackets (e.g., $\mathbf{x}[1]$ is the first component of the vector $\mathbf{x}$. For matrices, we use subscripts to identify entries ($W_{i,j}$ is the entry in row $i$, column $j$). Similarly, tensors are indexed using subscripts (e.g., $\mathcal{W}_{i,j,k}$ is the entry at row $i$, column $j$, depth $k$). To partially index a matrix or tensor, we use the symbol $*$. That is row $i$ in a matrix $W$ is $W_{i,*}$, column $j$ is $W_{*,j}$. Similarly, for a 4th order tensor $\mathcal{W}$, $\mathcal{W}_{i,j,*,*}$ is a matrix $V$ where $V_{k,\ell}=\mathcal{W}_{i,j,k,\ell}$.

Let
$D = \{(\textbf{x}_1, y_1), \ldots, (\textbf{x}_n, y_n)\}$ be a set of $n$ records,
where $\textbf{x}_i\in \mathfrak{X}$ is a feature vector and $y_i\in\mathfrak{Y}$ is a target (value we must learn to predict). We
say two datasets $D$ and $D^\prime$ are \emph{neighbors} if $D^\prime$
can be obtained from $D$ by adding or removing one record and write $D {\sim} D^\prime$ to denote this relationship.

\subsection{Differential Privacy}
Differential privacy is a widely accepted formal privacy definition that requires randomized algorithms (also called \emph{mechanisms} to process data. The intuition behind it is that the addition/deletion of one record should have very little influence on the output distribution.
\begin{definition}[($\epsilon,\delta$)-DP~\cite{Dwork2006calibrating,Dwork2006our}]
  \label{def:dp}%
Given privacy parameters $\epsilon\geq 0$, $\delta\geq 0$, a
randomized mechanism (algorithm) $\mathcal{M}$ satisfies ($\epsilon, 
\delta$)-differential privacy if for every set $S \subseteq
\Range(\mathcal{M})$ and for all pairs of neighboring datasets $D{\sim}D'$, 
\[
  \Pr[\mathcal{M}(D) \in S] \leq \exp(\epsilon)\Pr[\mathcal{M}(D')\in
  S] + \delta\,.
\]%
The probability only depends to the randomness in $\mathcal{M}$.
\end{definition}
The cases $\delta=0$ and $\delta>0$ are respectively referred to as  \emph{pure} and \emph{approximate} differential
privacy. 

%Pure differential privacy limits the ability of an attacker to make
%inferences about the specific record of any individual. Approximate
%differential privacy provides the same guarantees with probability
%$1-\delta$, and allows failure of privacy (e.g. release of the entire
%raw data) to occur with probability $\delta$. This event is known
%informally as the ``all-bets-are-off" scenario. 

%A recent relaxation of differential privacy, known as R\'enyi
%differential privacy \cite{Mironov2017renyi} provides weaker
%protections than pure differential privacy and stronger protections
%than approximate differential privacy (notably avoiding the
%``all-bets-are-off" scenario). 

\subsection{R\'enyi Differential Privacy}
One of the drawbacks of Definition \ref{def:dp} is that accurately tracking privacy loss from multiple noise-infused accesses to the data is difficult. For this reason, most work on differentially private deep learning uses a variant called R\'enyi Differential Privacy (RDP) \cite{Mironov2017renyi} to track privacy leakage of an iterative algorithm and then. At the very end, the RDP parameters are converted to the $\epsilon,\delta$ parameters of Definition \ref{def:dp}.
RDP relies on the concept of R\'enyi divergence:
%Pure differential privacy requires that constraint
%$e^{-\epsilon} \leq \frac{\Pr[\mathcal{M}(D) \in
%  S]}{\Pr[\mathcal{M}(D')\in S]} \leq e^{\epsilon}$ always holds, 
%R\'enyi differential privacy (RDP) \cite{Mironov2017renyi} allows this ratio to be a random
%variable and constrains it using the R\'enyi divergence: 

%is a recent relaxation of
%differential privacy, which unifies the notion of differential privacy 
%using $\alpha$-R\'enyi divergence.
%
\begin{definition}[R\'enyi Divergence]
Let $P_1$ and $P_2$ be probability distributions over a set $\Omega$ 
and let $\alpha\in (1,\infty)$. R\'enyi $\alpha$-divergence $\Divergence_\alpha$
is defined as:
$
%\[
  \Div{P_1}{P_2} = \frac{1}{\alpha-1}\log(\E_{x\sim P_2}\left[P_1(x)^\alpha P_2(x)^{-\alpha}\right])\,.
%\]
$
\end{definition}
R\'enyi differential privacy requires two parameters: a moment $\alpha$ and a parameter  $\epsilon$ that bounds the moment.
\begin{definition}[$(\alpha,\epsilon)$-RDP~\cite{Mironov2017renyi}]
Given a privacy parameter $\epsilon \geq 0$ and an $\alpha\in(1,\infty)$,
a randomized mechanism $\mathcal{M}$ satisfies $(\alpha,\epsilon)$-R\'enyi
differential privacy (RDP) if for all $D_1$ and $D_2$ that differ on the
value of one record,
%\[
$
\Div{\mathcal{M}(D_1)}{\mathcal{M}(D_2)} \leq \epsilon\,.
$
%\]
\end{definition}
%
%Note that when $\alpha=\infty$, R\'enyi divergence becomes max divergence and
%$(\alpha,\epsilon)$-RDP becomes pure $(\epsilon, 0)$-differential privacy.
While the semantics of RDP are still an area of research, its privacy guarantees are currently being interpreted in terms of $(\epsilon, \delta)$-differential privacy through the following conversion result \cite{Mironov2017renyi}.

\begin{lemma}[Conversion to $(\epsilon,\delta)$-DP~\cite{Mironov2017renyi}] \label{pro:conversion}
  If $\mathcal{M}$ satisfies $(\alpha, \epsilon)$-RDP, it satisfies $(\epsilon^\prime, \delta^\prime)$-differential privacy when $\epsilon^\prime \geq \epsilon + \frac{\log (1/\delta)}{\alpha-1}$ and $\delta^\prime \geq \delta$.
%  $(\epsilon + \frac{\log 1/\delta}{\alpha-1}, \delta)$-DP.
\end{lemma}

This result implies that $(\alpha,\epsilon)$-RDP can be converted to $(\epsilon^\prime, \delta^\prime)$-DP for many different choices of $\epsilon^\prime$ and $\delta^\prime$. The result can be used in many ways. For example, one may choose a desired $\epsilon^\prime$ and set $\delta^\prime=e^{-(\epsilon^\prime-\epsilon)(\alpha-1)}$, in which case $(\alpha, \epsilon)$-RDP provides more protections than differential privacy with those values of $\epsilon^\prime$ and $\delta^\prime$. Alternatively, one can pick a $\delta^\prime$ and use Lemma \ref{pro:conversion} to determine the corresponding $\epsilon^\prime$.

\vspace{0.5em}
\noindent\textbf{Building Blocks}.
One of the simplest methods of creating an algorithm satisfying RDP is called the Gaussian Mechanism. It relies on a concept called $L_2$ sensitivity, which measures the largest effect a single record can have on a function. Formally,
\begin{definition}[$L_2$ sensitivity]
Let $q$ be a vector-valued function over
datasets. The $L_2$ sensitivity of $q$, denoted by $\Delta_2(q)$ is
defined as   
%\[
$
  \Delta_2(q) = \max_{D{\sim}D'} \norm{q(D) - q(D')}_2\,,
$
%\]
where the max is over all  neighboring pairs.
\end{definition}

The Gaussian mechanism for RDP answers a numerical aggregate query $q$ by adding Gaussian noise whose variance depends on the sensitivity of $q$ as follows:

\begin{lemma}[Gaussian Mechanism~\cite{Mironov2017renyi}] \label{lem:rdp_gauss}
  Let $q$ be a vector-valued function over
datasets. Let $\mathcal{M}$ be a mechanism that releases the random variable
  $\mathcal{N}(q(D), \sigma^2\mat{I}_k)$ and let $\alpha\in (1,\infty)$ and $\epsilon >0$ be privacy parameters.
  %Then for any pair of
  %neighboring datasets $D$ and $D'$ and any $\alpha\in(1,\infty)$:
 % \[
%$
%\Div{\mathcal{M}(D)}{\mathcal{M}(D')}
%    \leq \alpha \Delta_2^2(q)/(2\sigma^2)\,.
%$
% \]%
%  In particular, 
If $\sigma^2 \geq \alpha\Delta_2^2(q)/(2\epsilon)$, then
  $\mathcal{M}$ satisfies $(\alpha, \epsilon)$-RDP.
\end{lemma}

\noindent\textbf{Composition.}
More complex algorithms for $(\alpha,\epsilon)$-RDP, such as training
deep neural networks, can be created by combining together many
applications of simpler mechanisms (such as the Gaussian Mechanism)
--- each one leaks a controlled amount of private information, and the
composition theorem explains how to compute the total leakage. 

\begin{lemma}[Composition \cite{Mironov2017renyi}]
Let $\mathcal{M}_1,\dots, \mathcal{M}_k$ be mechanisms such that each $\mathcal{M}_i$ satisfies $(\alpha, \epsilon_i)$-RDP (that is, the $\alpha$ values are all the same but the $\epsilon$ values can differ). The mechanism that, on input $D$, jointly releases the outputs $\mathcal{M}_1(D),\dots, \mathcal{M}_k(D)$ satisfies $(\alpha, \sum_i\epsilon_i)$-RDP.
\end{lemma}

In practice, one keeps track of multiple $\alpha$ values. That is, a mechanism $\mathcal{M}_1$ may satisfy $(\alpha_1, \epsilon_1)$-RDP, $(\alpha_2,\epsilon_2)$-RDP and $(\alpha_3,\epsilon_3)$-RDP, while $\mathcal{M}_2$ may satisfy  $(\alpha_1, \epsilon'_1)$-RDP, $(\alpha_2,\epsilon'_2)$-RDP and $(\alpha_3,\epsilon'_3)$-RDP. The mechanism that releases both of their outputs would satisfy $(\alpha_1, \epsilon_1+\epsilon'_1)$-RDP and also $(\alpha_2,\epsilon_2+\epsilon'_2)$-RDP and $(\alpha_3,\epsilon_3+\epsilon'_3)$-RDP. When converting  to $(\epsilon,\delta)$-DP, one applies Lemma \ref{pro:conversion} to each of these and selects the best $\epsilon,\delta$ values \cite{Abadi2016deep}.

\vspace{1em}
\noindent\textbf{Postprocessing Immunity.}
Another key feature of differential privacy  is post-processing immunity. If $\mathcal{M}$ is a mechanism that satisfy $(\alpha, \epsilon)$-RDP (or $(\epsilon',\delta')$-DP) and $f$ is any algorithm, then the mechanism which, on input $D$, releases $f(\mathcal{M}(D))$, satisfies $(\alpha, \epsilon)$-RDP (or $(\epsilon',\delta')$-DP) -- the privacy parameters do not get worse.

%% file: clipping.tex
In this section we briefly describe non-private training of neural networks to explain how GPU mini-batch computation is used to speed up training. We then discuss the most common differentially private deep learning training procedure and explain how its direct implementation loses much of these speed benefits (via a step called \emph{gradient clipping}). In Section \ref{sec:algorithm} we then explain how to recover the speedup that was lost with a better gradient clipping algorithm.

\subsection{Non-private mini-batch SGD.}

A machine learning model $M_{\vec{\theta}}$ is a parametrized function with parameters $\vec{\theta}$ (e.g., $\vec{\theta}$ could be the weights in an artificial neural network). Once the parameters are set, the model can make predictions. The parameters are typically chosen using training data through a process called \emph{empirical risk minimization}: given (1) a dataset $D=\{(\vec{x}_1, y_1), \dots, (\vec{x}_n, y_n)\}$ and (2) a loss function $\ell$ that quantifies the error between the true target $y_i$ and predicted value $M_\theta(\vec{x}_i)$, the goal of empirical risk minimization is to find a value of $\theta$ that minimizes
%
%Consider a machine learning model $M_{\theta}$ with a parameter vector
%$\vec{\theta} \in \mathbb{R}^k$ being trained on a labeled dataset
%$D=\{(\vec{x}_1, y_1), \ldots, (\vec{x}_n, y_n)\}$ where each
%$\vec{x}_i$ is a feature vector and $y_i$ is the target. The training
%process can be cast as a finite-sum minimization problem of form:
\begin{equation} \label{eq:finite_sum_loss}
  \underset{\vec{\theta} \in \Theta}{\argmin} L(\vec{\theta}, D)
  := \frac{1}{n}\sum_{i=1}^n \ell(y_i, M_{\vec{\theta}}(\vec{x}_i))\,.
\end{equation}
The function $L$ is called the \emph{objective function}.

 When $M_\theta$ is a deep neural network, the above problem is typically
solved with an iterative first-order algorithm such as stochastic
gradient descent (SGD)~\cite{Robbins1951stochastic,Bottou2010large} or
its variants. 

In each iteration, a set $B$ of $\tau$ records is randomly sampled from the
data $D$. This set is called a mini-batch. The objective function is computed over the minibatch: $L(\theta, B)=\frac{1}{\tau}\sum_{\vec{x}_\in B} \ell(y_i, M_{\vec{\theta}}(\vec{x}_i))$ and then its gradient $\nabla_{\vec{\theta}} L(\vec{\theta}, B)$ is computed. This gradient is then used to update the parameters $\vec{\theta}$, either through a vanilla update rule such as $\vec{\theta}\gets \vec{\theta} - \eta \nabla_{\vec{\theta}} L(\vec{\theta}, B)$ (where $\eta$ is a number called a \emph{learning rate}) or the gradient is used inside a more complicated procedure such as ADAM \cite{Kingma2014adam} or RMSProp (see \cite{RuderOpt} for a survey of alternatives).

The computation of the gradient $\nabla_{\vec{\theta}} L(\vec{\theta}, B)$ is generally the most expensive part of this procedure, but can be thought of as a series of matrix multiplications and element-wise products that can often be performed in parallel. Modern frameworks like TensorFlow \cite{tensorflow2015-whitepaper} and PyTorch \cite{Paszke2017automatic} use auto-differentiation (e.g., \texttt{torch.autograd.grad}) to compute the sum of the gradients over a batch (i.e., $\nabla_{\vec{\theta}} L(\vec{\theta}, B) \equiv \frac{1}{\tau}\sum_{\vec{x}_\in B}\nabla_{\vec{\theta}} \ell(y_i, M_{\vec{\theta}}(\vec{x}_i))$. Behind the scenes, data records are bulk-loaded onto the GPU to amortize data transfer costs and then the matrix operations take advantage of the parallelism in the GPU.

\subsection{Mini-batch stochastic gradient descent with privacy.}
In the framework of Abadi et al. \cite{Abadi2016deep}, adding differential privacy to deep learning requires adding bias and noise into the mini-batch gradient computation. Ideally, one would like to simply add noise to the minibatch gradient $\nabla_{\vec{\theta}} L(\vec{\theta}, B)\equiv \frac{1}{\tau}\sum_{\vec{x}_\in B}\nabla_{\vec{\theta}} \ell(y_i, M_{\vec{\theta}}(\vec{x}_i))$. \revision{To satisfy differential privacy, the noise has to be large enough to mask the effect of any possible record.} \revision{However, without any further assumptions, a worst-case change to a single record can result in a large change to the mini-batch gradient (potentially large enough to cause floating point computations to result in $\infty$)}. \revision{The amount of noise necessary to mask such an effect would render all computations useless.} \revision{Rescaling the inputs (e.g., converting image pixel values from the range [0,255] to [0,1]) would not solve this problem as the millions of weights in a deep network could still result in a large worst-case gradient (this happens even in the non-private setting and is called the exploding gradient problem \cite{GoodBengCour16}).}

\revision{Abadi et al. \cite{Abadi2016deep} addressed this problem by  clipping } each term in the summation to make sure that no term can get large, even in the worst case. The clipping function has a parameter $c$ (called the clipping threshold) and is defined as follows:
\begin{align*}
    \text{clip}_c(\vec{z}) = \frac{\vec{z}}{\max(1, || \vec{z}||_2/c)}
\end{align*}
If the $L_2$ norm of a vector is at most $c$, then $\text{clip}_c(\vec{z})=\vec{z}$ (and the $L_2$ norm of the result is $\leq c$). If the $L_2$ norm is $> c$ then $\text{clip}_c(\vec{z})=c\frac{\vec{z}}{||\vec{z}||_2}$, which has a norm equal to $c$. Hence $\text{clip}_c$ always outputs a vector of $L_2$ norm $\leq c$ that points in the same direction as the input vector.

Thus, the differentially private deep learning framework \cite{Abadi2016deep} replaces the mini-batch gradient with $\frac{1}{\tau}\sum_{\vec{x}_\in B}\text{clip}_c(\nabla_{\vec{\theta}} \ell(y_i, M_{\vec{\theta}}(\vec{x}_i)))$ and adds Gaussian noise (via the Gaussian mechanism) to this quantity before updating parameters in the network (e.g., the parameters can be updated with this noisy/biased gradient as in vanilla stochastic gradient descent: $\theta\gets \theta - \eta\left(\frac{1}{\tau}\sum_{\vec{x}_\in B}\text{clip}_c(\nabla_{\vec{\theta}} \ell(y_i, M_{\vec{\theta}}(\vec{x}_i)))\right)$ or the noisy/biased gradient can be used in more complex rules such as ADAM or RMSProp). Abadi et al. use the Moment Accountant technique \cite{Abadi2016deep} to precisely track the privacy protections offered by random sampling (to create the random mini-batches) and the added Gaussian noise.

\subsection{The Computational Problem}
The efficiency of the differentially private deep learning framework depends on the following question: how does one compute $\frac{1}{\tau}\sum_{\vec{x}_\in B}\text{clip}_c(\nabla_{\vec{\theta}} \ell(y_i, M_{\vec{\theta}}(\vec{x}_i)))$?
Auto-differentiation software will not do this directly.

One baseline approach (as implemented in TensorFlow Privacy \cite{tensorflowprivacy}) is to loop through the examples one at a time. For each example $\vec{x}_i$ one can ask the auto-differentiator to compute $\nabla_{\vec{\theta}} \ell(y_i, M_{\vec{\theta}}(\vec{x}_i))$, then clip it and then at the end, sum up the clipped gradients.

This approach has several drawbacks. First, it loses the parallelism that GPUs can offer when performing matrix computations. Second, it may result in multiple transfers of data to the GPU (i.e., not taking advantage of bulk transfer capabilities).

A related,  slightly faster approach is to use the auto-differentiation api to directly ask for multiple gradients. For example in PyTorch, the function \texttt{torch.autograd.grad} is normally called with the first parameter equal to the minibatch loss $\frac{1}{\tau}\sum_{\vec{x}_\in B} \ell(y_i, M_{\vec{\theta}}(\vec{x}_i))$ (in which case it computes the gradient). However, it is also possible to call the function with a vector of losses: $[\ell(y_1, M_{\vec{\theta}}(\vec{x}_1)), \dots, \ell(y_\tau, M_{\vec{\theta}}(\vec{x}_\tau))]$ to obtain the gradient of each one. These gradients can then be clipped and summed together.

In our experiments, this approach is still significantly slower than non-private training. Further significant improvements are possible and are described in Section \ref{sec:algorithm}. The main idea is that when deep learning auto-differentiators compute the gradients, they are also computing the derivatives with respect to intermediate variables (e.g., the chain rule). Normally, these intermediate results are not returned but it is possible to ask for them. The per-example gradient norms (i.e. norm of the gradient of each term $\ell(y_i, M_{\vec{\theta}}(\vec{x}_i)$) can be directly computed from these intermediate results. Once the per-example gradient norms are computed, we turn them into weights $\nu_1,\dots, \nu_\tau$ then re-weight the terms in the mini-batch loss: $\frac{1}{\tau}\sum_{\vec{x}_\in B}\;\nu_i \;\ell(y_i, M_{\vec{\theta}}(\vec{x}_i))$. This step ensures that the gradient of each weighted term now has norm at most $c$. We then ask the audo-differentiator for the gradient of this reweighted loss. The result is exactly equivalent to per-example gradient clipping (but turns out to be much faster than the baseline implementations). Thus, after this reweighted gradient is computed, noise can be added and parameters can be updated as in \cite{Abadi2016deep}. We describe the details in Section \ref{sec:algorithm}.

%% file: related.tex
% There has been extensive research on differentially private optimization
% algorithms for machine learning
% problems~\cite{Song13sgd,Zhang2013privgene,Wang2015free,Lee2018cdp,Iyengar2019amp},
% especially for empirical risk minimization
% (ERM)~\cite{Chaudhuri2011Objpert,Kifer2012erm,Bassily2014PERM,Zhang2017RR,Wang2017dpsvrg,Chen2019renyi}   
% and deep neural networks
% (DNN)~\cite{Abadi2016deep,Papernot2018pate,Jordon2019pategan,Yu2019modelpub}. For 
% nonconvex problems such as DNNs, gradient perturbation-based
% algorithms~\cite{Song13sgd,Bassily2014PERM,Abadi2016deep,Lee2018cdp,Yu2019modelpub} 
% are widely used due to their simplicity and less restrictive requirement
% on the objective function.

Deep learning for differential privacy was introduced by Skokri and Shmatikov \cite{RezaDeep} but required enormous values of the privacy parameters (e.g., $\epsilon$ values in the hundreds or thousands). The first practical approach, which could train deep networks to reasonable accuracy (on the MNIST and CIFAR datasets) with $\epsilon$ values of 10 or less was proposed by Abadi et al. \cite{Abadi2016deep} and required the use of gradient clipping and Renyi Differential Privacy \cite{Mironov2017renyi} (referred to as the Moment Accountant in  \cite{Abadi2016deep}).

Followup work  \cite{Yu2019modelpub,AcsGAN,thakkarclip,Brendan2018learning,ChenDPGAN,BeaulieuJones159756,Abay2018:PPS} relied on this training technique. Also \cite{Yu2019modelpub,AcsGAN,thakkarclip,Brendan2018learning} investigated different clipping strategies, such as adaptively changing the clipping threshold \cite{Yu2019modelpub,thakkarclip,AcsGAN} or clipping the gradient layer by layer \cite{Brendan2018learning,Mcmahan2018general}. Specifically, given the global clipping threshold $c$,
McMahan et al.~\cite{Brendan2018learning} clip the gradient of each layer's parameter using the threshold
$c/\sqrt{m}$, where $m$ is the total number of layers. 
%Since our
%proposed fast per-example clipping framework is able to compute the
%per-example gradient norm layer-wise, our work can be used to
%accelerate their training algorithm. 
In~\cite{Mcmahan2018general}, the authors extended the idea of
per-layer clipping and proposed a joint clipping strategy which
applies different amount of clipping to each group of queries.
Since our
proposed fast per-example clipping framework is able to compute the
per-example gradient norm layer-wise (as well as overall norm), our work can be used to
accelerate the previously mentioned training algorithms that experimented with more refined clipping ideas.

There are other approaches to differentially private training of deep networks that avoid gradient clipping and adding noise to gradients. One example is PATE \cite{pate,Papernot2018pate} which requires a large private dataset but also a large public dataset (and hence is applicable in fewer scenarios). Gradient clipping in specific models can also be avoided, for example Phan et al. \cite{PhanAAAI2016} perturb the objective function of auto-encoders while  Xie et al. \cite{Xie:DPGAN} show that it is possible to train a differentially private GAN using weight clipping instead of gradient clipping.

%Gradient clipping~\cite{Abadi2016deep} is an essential technique
%widely used in differentially private machine 
%learning to bound a single user's contribution to the query result
%(i.e., the gradient). 
%It is particularly popular when training a neural network as there
%exists no a priori known bound on its
%gradient~\cite{Brendan2018learning,xie2018differentially,Yu2019modelpub}. 
%
%McMahan et al.~\cite{Brendan2018learning} used the gradient clipping
%to build an LSTM language model in a federated learning setting. They
%introduced \emph{per-layer clipping} strategy in which the learning
%algorithm clips the gradient of each layer's parameter separately,
%rather than clipping the concatenation of all
%parameters. Specifically, given the global clipping threshold $C$,
%they clip the gradient of each layer's parameter using the threshold
%$C/\sqrt{m}$, where $m$ is the total number of layers. Since our
%proposed fast per-example clipping framework is able to compute the
%per-example gradient norm layer-wise, our work can be used to
%accelerate their training algorithm. 
%In~\cite{Mcmahan2018general}, the authors extended the idea of
%per-layer clipping and proposed a joint clipping strategy which
%applies different amount of clipping to each group of queries.
%
%To bound the sensitivity of gradients, Xie et
%al.~\cite{xie2018differentially} applies the idea of clipping to the
%network's weight vectors rather than to the gradients.

Overall, basing differentially private training algorithms on gradient clipping techniques (e.g., \cite{Abadi2016deep})
results in algorithms that are applicable in wider settings.
However, despite the popularity of gradient clipping technique in
differentially private deep learning, per-example gradient computation
for a general neural network was
computationally heavy and significantly slowed down training.

In~\cite{Goodfellow2015efficient}, Goodfellow showed that for
fully-connected networks per-example gradients can be efficiently
computed using auto-differentiation library in deep learning
frameworks, such as Tensorflow and PyTorch. A key observation is that in these specific networks,
the gradient of loss function $L$, defined
in~\eqref{eq:finite_sum_loss}, with respect to the network parameters 
can be decomposed into the product of intermediate results of the
auto-differentiation procedure. Specifically, consider a fully-connected layer
with, weight matrix $W \in \R^{m\times n}$ and bias $\vec{b} \in
\R^m$,
whose pre-activation $\vec{z} \in \R^m$ are computed by
%\[
$
  \vec{z} = W\vec{h} + \vec{b}\,,
$
%\]
where $\vec{h} \in \R^n$ is an input vector to the layer (or
equivalently, it is the post-activation of the previous layer).
A careful analysis using the chain rule reveals that
\[
  \norm*{\pdv{L}{W}}_{\mathsf{F}}^2 = \norm*{\pdv{L}{\vec{z}}}^2 \norm{\vec{h}}^2\,.
\]
Hence, the norms of per-example gradients can be efficienlty computed
(without having to explictly materialize them) if
we store $\vec{z}$ and $\vec{h}$ and  compute $\pdv{L}{\vec{z}}$ using
the auto-differentiation library. However, this formula does
\emph{not} generalize to other type of neural network
layers, e.g., convolutional layer and recurrent layer. We observe that
the technique is applicable when the 
gradient with respect to parameter is expressed as an outer product
between the gradient with respect to pre-activation $\pdv{L}{\vec{z}}$ and
the layer input $\vec{h}$. That is when
\[
  \pdv{L}{W} = \pdv{L}{\vec{z}} \otimes \vec{h}\,,
\]
where $\otimes$ denotes the outer product of two vectors.
In this work, we extend the technique to other types of neural networks,
derive equations for per-example gradients, and provide a  recipe for
efficiently computing them and integrating them into differentially
private training.

Recently, at the time of writing, Rochette et
al.~\cite{Rochette2019EfficientPG} also made an attempt to extend the
technique in~\cite{Goodfellow2015efficient} to convolutional neural
networks. While they also analyzed gradients using the chain rule and
made observations similar to those in our work,
their work differs with ours in both mathematical derivation and
implementation. For simplicity, \cite{Rochette2019EfficientPG}
derives the gradient for 1D convolution operation and claim the same
result also holds for higher dimensional cases. In our work, we
directly show the derivations for 2D convolution (which is most
popularly used in practice) using tensors. 
Another aspect of their technique is that to compute the per-example gradients
for 1-D convolutions, they make use of 2-D convolution operations. Extensions of their
techniques to per-example gradients for 2-D convolutions would require 3-D convolutions
and extensions of their work to 3-D convolutions would not be efficiently supported (for example, 
due to lack of efficient support of 4-D convolutions in PyTorch).
%
%In their implementation of
%per-example gradient computation, they decompose the gradient
%computation into multiple 2D convolution operations being separately
%applied to the same image and concatenate the results. The benefit of
%this is that it can be done using the existing \textsf{conv2d}
%function with simple reshaping and grouping. 
In contrast, to avoid this problem in our
implementation, we convert the same operation into one single batch
matrix-matrix multiplication, which can be done efficiently on
GPUs. In addition, to smoothly integrating our technique into differentially
private training, we indirectly clip gradients by assigning weights to
loss values, rather than directly manipulating the gradients.

%%% Local Variables:
%%% mode: latex
%%% TeX-master: "main"
%%% End:

%% file: algorithm.tex
In this paper, we consider feedforward networks (which include recurrent networks) consisting of layers (e.g., a convolutional layer feeding into a max pooling layer, etc.).

\begin{algorithm}[t]  
  \DontPrintSemicolon
  \KwIn{dataset $D=\{d_i\}$, model $M_{\vec{\theta}}$, activation
    function $\phi(\cdot)$, privacy parameters $\epsilon$ and $\delta$, number of iterations $T$, mini batch size $\tau$, clipping threshold $c$}
    Use Moment Accountant \cite{Abadi2016deep} to determine noise variance $\sigma^2$ (based on $m$, $c$, and $T$) that will result in $(\epsilon, \delta)$-dp.\;
  \For{$t=1,2,\ldots, T$}{%
    Construct a random minibatch $B$ of $\tau$ records\;
    $\Gamma = \emptyset\,, \Lambda=\emptyset$\;
    \tcc{Perform the feed forward step}
    \ForEach{\text{layer} $l$ \text{in} $M_{\vec{\theta}}$}{%
      $Z^{(l)} = X^{(l-1)}W^{(l)} + \vec{b}^{(l)}$\;
      $X^{(l)} = \phi(Z^{(l)})$\;
      $\Gamma = \Gamma \cup \{Z^{(l)}\}$ \tcp*{pre-activation}
      $\Lambda = \Lambda \cup \{X^{(l-1)}\}$\tcp*{layer input}
    }
    Compute $\pdv{L(\vec{\theta}, B)}{\Gamma}$ via auto-differentiation\;
    \tcp*{same as \pdv{\frac{1}{\tau}\sum_{\vec{x}_\in B} \ell(y_i, M_{\vec{\theta}}(\vec{x}_i))}{\Gamma}} \label{line:autograd}
    Using $\Lambda$ and $\pdv{L(\vec{\theta}, B)}{\Gamma}$, compute $||\grad{\ell}(y_i, M_{\vec{\theta}}(\vec{x}_i))||_2$ for $i\in B$ as described in Section \ref{sec:algorithm}\; \label{line:pe_grad}
    $\nu_i \gets \min(1,\, c/\norm{\grad{\ell}(\vec{\theta}, d_i)}_2)$\;
    
    Use auto-differentiation to compute gradient of: $\frac{1}{\tau}\sum_{\vec{x}_\in B} \nu_i\ell(y_i, M_{\vec{\theta}}(\vec{x}_i))$\;
    \tcp*{Add Gaussian Noise to the gradient (as in \cite{Abadi2016deep}) and update parameters.}
    $\theta\gets \theta - \eta \left(N(0, \sigma^2I) _ +\frac{1}{\tau}\nabla_{\vec{\theta}}\sum_{\vec{x}_\in B} ~\nu_i  \ell(y_i, M_{\vec{\theta}}(\vec{x}_i))\right) $
  }
  \caption{Gradient perturbation by reweighting per-example losses \label{alg:reweight}}  
\end{algorithm}

Each layer $\ell$ has a weight matrix $W^{(\ell)} \in \R^{d_{\mathrm{in}}\times d_{\mathrm{out}}}$ where $d_{\mathrm{in}}$ is the number of inputs to the layer and $d_{\mathrm{out}}$ is the number of outputs.

Since each example in the mini-batch is being run through the network, we can think of the inputs to the layer as a matrix $X^{(\ell)} \in \R^{\tau\times d_{\mathrm{in}}}$ whose first row (i.e., $X^{(\ell)}_{1,*}$) is the layer's input when the first record of the mini-batch is run through the network and the $i^\text{th}$ row (i.e., $X^{(\ell)}_{i, *}$) is the layer's input when the $i^\text{th}$ record of the mini-batch is run through the network.

The pre-activation of the layer is then $X^{(\ell)}W^{(\ell)}+\vec{b}$, where $\vec{b}^{(\ell)}\in\R^{d_{\mathrm{out}}}$ is the bias parameter of the layer.

The activation function $\phi^{(\ell)}$ of the layer is applied pointwise to the pre-activation to give the post-activation, or output, of the network:

\[
  X^{(\ell+1)} = \phi^{(\ell)}(Z^{(\ell)})\,,\quad Z^{(\ell)} = X^{(\ell)}W^{(\ell)} + \vec{b}^{(\ell)}\,,
\]

In this section we show how to compute $\frac{1}{\tau}\sum_{\vec{x}_\in B}\text{clip}_c(\nabla_{\vec{\theta}} \ell(y_i, M_{\vec{\theta}}(\vec{x}_i)))$. This is the quantity to which Gaussian noise is added and which is then used to update the network parameters during training. Pseudocode for the integration of our procedure into differentially private deep learning in shown in Algorithm \ref{alg:reweight}.

The main idea behind our approach is that $\text{clip}_c(\nabla_{\theta}\ell(y_i, M_{\vec{\theta}}(\vec{x}_i)))=\nu_i \nabla_{\theta}\ell(y_i, M_{\vec{\theta}}(\vec{x}_i))$, where 
\begin{equation}
\nu_i=\min(1, c/||\nabla_{\theta}\ell(y_i, M_{\vec{\theta}}(\vec{x}_i))||_2)\label{eq:per_example_weight}
\end{equation}

If we can compute $\nu_i$ for each $\vec{x}_i$, then the  re-weighted loss on the minibatch:
\begin{equation}
\frac{1}{\tau}\sum_{\vec{x}_\in B}\;\nu_i \;\ell(y_i, M_{\vec{\theta}}(\vec{x}_i))\label{eq:weighted_loss}
\end{equation}
has gradient that equals $\frac{1}{\tau}\sum_{\vec{x}_\in B}\text{clip}_c(\nabla_{\vec{\theta}} \ell(y_i, M_{\vec{\theta}}(\vec{x}_i)))$.

Thus we compute $\nu_i$ for each $i$, reweight the loss function, ask the auto-differentiation api to get the gradient, add privacy noise to the gradient, and then update the parameters. The result is exactly the same as per-example gradient clipping, but is much faster.

Noting that the parameters $\vec{\theta}$ consists of the weight matrix and bias vector of each layer, the $L_2$ norm of the gradient with respect to $\vec{\theta}$ is the square root of the sum of squares of the gradients with respect to the $W^{(\ell)}$ and $\vec{b}^{(\ell)}$ of each layer.

Thus, in each of the following subsections, we explain 
how to compute these quantities for each type of layer.
All that is needed are quantities  $\pdv{\text{Loss}}{Z^{(\ell)}}$ (the gradient with respect to pre-activations of Layer $\ell$) and $X^{(\ell)}$ (the mini-batch inputs to Layer $\ell$).

\subsection{Fully-connected Layers}
For completeness, we first describe Goodfellow's technique for fully connected layers \cite{Goodfellow2015efficient}.

\begin{figure}[tp]
  \centering
  \begin{tikzpicture}[
    myunit/.style={circle,draw,fill=gray!50,minimum width=1em,minimum height=1em,on chain},
    mylayer/.pic={%
      \begin{scope}[start chain=going right, node distance=3pt]
        \foreach \i in {1,...,#1}{%
          \node[myunit] (-x\i) {};
        }
        \node[fit={(-x1)(-x#1)},draw,rounded corners=3pt] (-bb) {};
      \end{scope}}]
    \pic[local bounding box=l0] (l0) at (0, 0) {mylayer={5}};
    \pic[local bounding box=l1] (l1) at (0, 1.5) {mylayer={5}};
    \node[draw=none,right=.5cm of l0-bb] (eq2) {$\vec{x} \in \R^m$};
    \node[draw=none,left=.3cm of l0-bb] (label2) {layer $\ell-1$};
    \node[draw=none,right=.5cm of l1-bb] (eq1) {$\vec{z}= W\vec{x} + \vec{b} \in \R^n$};    
    \node[draw=none,left=.3cm of l1-bb] (label1) {layer $\ell$};
    \draw[-latex,thick,draw=gray] (l0-bb.north) -- (l1-bb.south) node[right=5pt,midway] {$W\in \R^{m\times n}$};
\end{tikzpicture}
\caption{Two fully-connected layers in an MLP}
\label{fig:fully_connected_layer}
\end{figure}
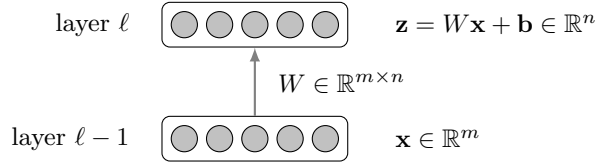

Consider two consecutive fully-connected layers, described in 
Figure~\ref{fig:fully_connected_layer}, of a multi-layer perceptron 
(MLP). Let $\ell$ and $\ell-1$ denote those two layers. Let $W \in
\R^{m\times n}$ be the weight matrix between layers $\ell$ and $\ell-1$
and $\vec{x} \in \R^{n}$ be an input to the upper
layer (which is also the activation of bottom layer). In the forward phase, the
pre-activation $\vec{z}\in \R^m$ and activation $\vec{a} \in \R^m$ are: 
\begin{equation} \label{eq:mlp_preact} 
  \vec{z} = W\vec{x} + \vec{b}\,,\qquad  \vec{a} = \phi(\vec{z})\,,
\end{equation}
where $\phi$ is an activation function applied element-wise and $\vec{b} \in
\R^m$ is a bias term.
By the chain rule, the derivative of $L$
with respect to the entry of $W$ at $i$\th row and $j$\th column is given by
\begin{align}
  \pdv{L}{W_{i,j}}
  &= \pdv{L}{\vec{z}} \pdv{\vec{z}}{W_{i,j}} = \sum_{k=1}^m
    \pdv{L}{\vec{z}[k]}\pdv{\vec{z}[k]}{W_{i,j}} \,, \label{eq:mlp_dldw_chain}
\end{align}
where we view $\pdv{L}{\vec{z}}$ and $\pdv{\vec{z}}{W_{i,j}}$ as
matrices of size $1\times m$ and $m\times 1$, respectively. From~\eqref{eq:mlp_preact}
we have  
\begin{align*}
  \pdv{\vec{z}[k]}{W_{i,j}}
  &=\pdv{}{W_{i,j}}\left(\sum_{l=1}^n W_{k,l} \vec{x}[l] + \vec{b}[k]\right) 
  =
    \begin{cases}
      \vec{x}[j] & \mbox{ if $k=i$,}\\
      0 & \mbox{ if $k\neq i$.}
  \end{cases}
\end{align*}
Plugging the above into~\eqref{eq:mlp_dldw_chain}, we obtain
\[
  \pdv{L}{W_{i,j}} = \pdv{L}{\vec{z}[i]} \vec{x}[j]\,.
\]
Combining all together, we see that
\begin{align}
  \pdv{L}{W}
  &=
    \begin{bmatrix}
      \pdv{L}{\vec{z}[1]} \vec{x}[1] & \pdv{L}{\vec{z}[1]} \vec{x}[2] & \cdots & \pdv{L}{\vec{z}[1]}\vec{x}[n]\\[1ex]
      \pdv{L}{\vec{z}[2]} \vec{x}[1] & \pdv{L}{\vec{z}[2]} \vec{x}[2] & \cdots & \pdv{L}{\vec{z}[2]} \vec{x}[n]\\[1ex]
      \vdots & \vdots & \ddots & \vdots \\[1ex]
      \pdv{L}{\vec{z}[m]} \vec{x}[1] & \pdv{L}{\vec{z}[m]} \vec{x}[2] & \cdots & \pdv{L}{\vec{z}[m]} \vec{x}[n]\\
    \end{bmatrix} \nonumber \\
  &= \pdv{L}{\vec{z}} \otimes \vec{x}\,, \label{eq:grad_linear_layer}
\end{align}
where $\otimes$ denotes the outer product of two
vectors. Equation~\eqref{eq:grad_linear_layer} is the gradient for a
single example $\vec{x}$. Suppose there are $\tau$ examples in the
minibatch. Then both $\pdv{L}{\vec{z}}$ and $\vec{x}$ become matrices
  of size $\tau\times m$ and $\tau\times n$, respectively. To
  efficiently compute the per-example graidents for $\tau$ examples,
  in our implementation, we reshape $\pdv{L}{\vec{z}}$ and $\vec{x}$
  into tensors of size $[\tau, m, 1]$ and $[\tau, 1, n]$,
  respectively, and perform batch matrix-matrix multiplication\footnote{In
    PyTorch, this is done using \texttt{torch.bmm()} function.}. This
  procedure is described in
  Algorithm~\ref{alg:pegrad_fc_layer}. We note that in the pseudocode
  $\pdv{L}{\vec{z}_i}$  denotes the gradient for the $i$\th example in
  the minibatch.  
\begin{algorithm}[tp]  
  \DontPrintSemicolon
  \KwIn{batch of gradients w.r.t. pre-activations $Z =
    [\pdv{L}{\vec{z}_1}^\intercal, \ldots,\pdv{L}{\vec{z}_{\tau}}^\intercal]^\intercal$,
    batch of layer's input $X=[\vec{x}_1^\intercal, \ldots, \vec{x}_{\tau}^\intercal]^\intercal$}
  $\mathcal{Z} \gets$ reshape $Z$ into $[\tau, m, 1]$\;
  $\mathcal{X} \gets$ reshape $X$ into $[\tau, 1, n]$\;
  \tcc{Compute batch matrix-matrix multiplication}
  $\mathcal{G}$ = \textsf{bmm}($\mathcal{Z}$, $\mathcal{X}$)\;
  \Return $\mathcal{G}$\;
  \caption{Per-example gradient computation for fully-connected layer}
  \label{alg:pegrad_fc_layer}
\end{algorithm}
Similarly, the gradient of $L$ with respect to the $k$\th entry of bias term $\vec{b}[k]$ is
\begin{align*}
  \pdv{L}{\vec{b}[k]}
  &= \pdv{L}{\vec{z}}\pdv{\vec{z}}{\vec{b}[k]} = \pdv{L}{\vec{z}} I_m = \pdv{L}{\vec{z}}\\
  \intertext{since we have}
  \pdv{\vec{z}[p]}{\vec{b}[k]}
  &= \pdv{}{\vec{b}[k]}\left(\sum_{l=1}^m W_{p,l}\vec{x}[l] + \vec{b}[p]\right) 
  =\begin{cases}
      1 & \mbox{ if $p=k$,}\\
      0 & \mbox{ if $p\neq k$,}\\
    \end{cases}
\end{align*}
where $I_m$ denotes an identity matrix of size $m\times m$.

\subsection{Convolutional Layers}
Suppose we have a convolutional layer with $c_{\text{out}}$ kernels of
size $\kappa\times \kappa$~\footnote{Here we assume the width and height of
  filter are the same for simplicity. Our result can be generalized to
  the filters with abitrary size.}. Assume input images have size
$s_H\times s_W$ with $c_{\text{in}}$ channels. The kernel $\mathcal{W}$ for
the layer can be represented by a 4D tensor with dimensions
$[c_{\text{out}}, c_{\text{in}}, \kappa, \kappa]$, and the input image
$\mathcal{X}$ by a 3D tensor with dimensions $[c_{\text{in}}, s_H, s_W]$. We denote the
entry of tensor $\mathcal{X}$ at location $(i, j, k)$ by
$\mathcal{X}_{i,j,k}$ and write 
$\mathcal{X}_{i, j, *}$ to denote the entries of $\mathcal{X}$ whose indices for the first
2 dimensions are fixed to $(i, j)$. $\mathcal{X}_{i:j}$ denotes the entries with
indices from $i$ to $j$.

\begin{figure}[tp]
  \centering
  \def\imageh{3.5}
  \def\outimgh{2.5}
  \def\nchannel{0.7}
  \def\outchannel{1.0}
  \def\koffset{0.25}
  \def\kernelh{1.5}
  \begin{tikzpicture}[innerk/.style={dotted,thick,gray},
    pics/myimg/.style n args={6}{code={%
        \draw[#6] (#1, 0, 0) --node[below,midway] {#2} ++  (-#1, 0, 0);
        \draw[#6] (#1, 0, 0) --node[right,midway] {#3} ++(0,0,-#5) --++ (0,#5,0);
        \draw[#6] (#1, 0, 0) -- ++(0,#5, 0) --++ (0,0,-#5);
        \draw[#6] (0, 0, 0) --node[left,midway] {#4} (0, #5, 0) -- (0, #5, -#5) -- (#1, #5, -#5);
        \draw[#6] (0, #5, 0) -- (#1,#5, 0);
      }}]
    % input image
    \pic[local bounding box=l0] (l0) at (0, 0, 0)
    {myimg={\nchannel}{$c_{\mathrm{in}}$}{$s_W$}{$s_H$}{\imageh}{draw=black}};
    % kernel
    \pic[local bounding box=l2] (l2) at (0, \imageh*\koffset, -\imageh*\koffset)
    {myimg={\nchannel}{}{}{}{\kernelh}{draw=gray,dotted,thick}}; 
    % \pic[local bounding box=l1] (l1) at (0, 0, 0) {kernel={dotted,thick,gray}};
    % kernel
    \pic[local bounding box=l2] (l2) at (2.1, \imageh*\koffset, -\imageh*\koffset)
    {myimg={\nchannel}{$c_{\mathrm{in}}$}{$\kappa$}{$\kappa$}{\kernelh}{draw=red}};    
    % output
    \coordinate (out) at (4.5, {(\imageh-\outimgh)/2}, -{(\imageh-\outimgh)/2});
    \draw[thick,fill=gray!20] ($(out)+(0.2, 0, 0)$) --++
    (0, \outimgh, 0) --++ (0, 0, -\outimgh) --++ (0, -\outimgh, 0) --cycle;  
    \pic[local bounding box=l3] (l3) at (out)
    {myimg={\outchannel}{$c_{\mathrm{out}}$}{$d_W$}{$d_H$}{\outimgh}{draw=black}};
    \coordinate (tc) at ($(out)+(0.2,\outimgh,-\outimgh/2)$);
    \node[draw=none] (fmap) at ($(tc)+(-0.8, 0.8, 0)$) {$\mathcal{Z}_{l,*,*}$};
    \draw[-latex] (fmap) edge[bend left] (tc);
    \node[draw=none,below=5pt of l2] (wlabel) {$\mathcal{W}_{l,*,*,*}$};
    \node[draw=none,below=5pt of l3] (zlabel) {$\mathcal{Z}$};
    \node[draw=none,above left=-15pt and -25pt of l0.north west] (xlabel) {$\mathcal{X}$};
  \end{tikzpicture}
  \caption{Convolution between $\mathcal{W}$ and $\mathcal{X}$}
  \label{fig:convolutional_layer}
\end{figure}
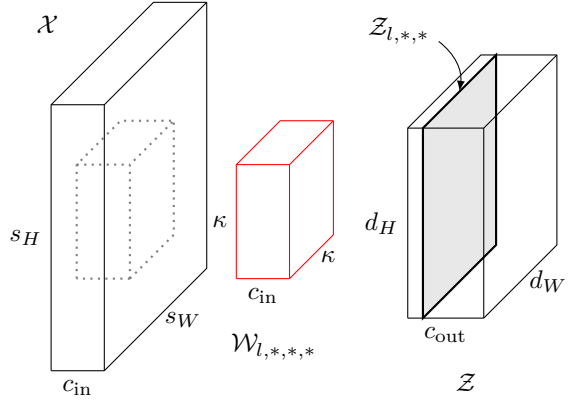

The pre-activation $\mathcal{Z}$ resulting from performing convolution between
$\mathcal{W}$ and $\mathcal{X}$, denoted by $\mathcal{W}*\mathcal{X}$, is expressed as 
\begin{align}
  \mathcal{Z}_{l, m, n}
  &=  \mathcal{W}_{l, *, *, *} \star \mathcal{X}_{*, m:m+\kappa, n:n+\kappa} + b_l\,, \label{eq:conv_preact}
\end{align}
where $\star$ symbol defines the inner product between two tensors of same
order, i.e., $\mathcal{X}\star \mathcal{Y} = \sum_{i,j,k}
\mathcal{X}_{i,j,k}\mathcal{Y}_{i,j,k}$.
For simplicity, let's fix $l$ and focus on the $l$\th output feature
map. See Figure~\ref{fig:convolutional_layer} for a graphical
depiction of the convolution operation.
From~\eqref{eq:conv_preact}, we get
\begin{align*}
  \pdv{\mathcal{Z}_{l,m,n}}{\mathcal{W}_{l,k,i,j}}
  &=\pdv{\sum_{p=1}^{\kappa}\sum_{q=1}^{\kappa}\sum_{r=1}^{c_{\text{in}}}\mathcal{W}_{l,r, p, q}\mathcal{X}_{r,m+p{-}1,n+q{-}1}}{\mathcal{W}_{l,k,i,j}}\\
  &=\mathcal{X}_{k, m+i-1, n+j-1}\,
\end{align*}
and see that the derivative of the $l$\th pre-activation
with respect to $\mathcal{W}_{l,k,i,j}$ is given by 
\begin{align*}
  &\pdv{\mathcal{Z}_{l,*,*}}{\mathcal{W}_{l,k,i,j}}\\
  &=
    \begin{bmatrix}
      \mathcal{X}_{k,i,j} & \mathcal{X}_{k,i,j+1} & \cdots & \mathcal{X}_{k,i,j+d_W}\\
      \mathcal{X}_{k,i+1,j} & \mathcal{X}_{k,i,j+1} & \cdots & \mathcal{X}_{k,i+1,j+d_W}\\
      \vdots & \vdots & \ddots & \vdots \\
      \mathcal{X}_{k,i+d_H,j} & \mathcal{X}_{k,i+d_H,j+1} & \cdots & \mathcal{X}_{k,i+d_H,j+d_W}\\
    \end{bmatrix}\,,
\end{align*}
where $d_H = s_H-\kappa$ and $d_W=s_W-\kappa$.
Using the chain rule, we get
\begin{equation} \label{eq:grad_conv2d_derivation}
\begin{aligned}
  \pdv{L}{\mathcal{W}_{l,k,i,j}}
  &= \pdv{L}{\mathcal{Z}_{l,*,*}} \pdv{\mathcal{Z}_{l,*,*}}{\mathcal{W}_{l,k,i,j}} \\
  &= \sum_{m'=1}^{d_H+1}\sum_{n'=1}^{d_W+1} \pdv{L}{\mathcal{Z}_{l,m',n'}}\mathcal{X}_{k,i+m'-1,j+n'-1}\\
  &= \pdv{L}{\mathcal{Z}_{l,*,*}} \star \mathcal{X}_{k,i:i+d_H,j:j+d_W}\,.
\end{aligned}
\end{equation}
The above equation implies that the gradient $\pdv{L}{\mathcal{W}_{l,k, *,*}}$
is obtained by performing convolution between the derivative of $L$
with respect to the pre-activation $\mathcal{Z}_{l,*,*}$ and input image $\mathcal{X}_{k,*,*}$
(without the bias term). That is,
\[
  \pdv{L}{\mathcal{W}_{l,k,*,*}}
  = \left(\pdv{L}{\mathcal{Z}}\right)_{l,*,*} * \mathcal{X}_{k,*,*}\,.
\]
As described for the fully-connected layer case, the per-example
gradient can be obtained from the derivative $\pdv{L}{\mathcal{Z}}$ and layer's
input $\mathcal{X}$. The only difference is that we now need to compute the
convolution between these two tensors --- it was outer product in the
fully-connected layer case.
To efficiently perform the above convolution operation, we convert it
into a general matrix-matrix multiplication
(GEMM)~\cite{Chellapilla2006high} through vectorizing 
and reshaping the data and leverage its fast implementation in BLAS
library. To this end, we apply \textsf{im2col}~\cite{Jia2014learning}
transformation on images which converts an image into a matrix where
each row corresponds to $\kappa\times \kappa\times C_{\mathrm{in}}$
pixels to which the kernel is applied. See
Algorithm~\ref{alg:pegrad_conv_layer} for the procedure to get
per-example graidents using this operation.

\noindent\textbf{Extensions to 3D convolution}. The derivation
in~\eqref{eq:grad_conv2d_derivation} readily generalizes to 3D
case. Consider a 3D convolution between an input $\mathcal{X}$ of
shape $[c_{\mathrm{in}}, d_{\mathrm{in}}, s_H, s_W]$ and a kernel
$\mathcal{W}$ of shape $[c_{\mathrm{out}}, c_{\mathrm{in}},
\kappa,\kappa, \kappa]$. The entry at position $(o, m, n)$ of the
$l$\th output feature map is given by
\[
  \mathcal{Z}_{l,o,m,n}=\sum_{p=1}^\kappa\sum_{q=1}^\kappa\sum_{r=1}^{\kappa}\sum_{c=1}^{c_{\mathrm{in}}}
  \mathcal{W}_{l,c,r,p,q}\cdot X_{c,o+r,\,m+p-1,n+q-1}\,.
\]
From the above, it is easy to see that
\[
  \pdv{\mathcal{Z}_{l,o,m,n}}{\mathcal{W}_{l,c,k,i,j}} = \mathcal{X}_{c,o+k-1,m+i-1,n+j-1}
\]
and $\pdv{\mathcal{Z}_{l,*,*,*}}{\mathcal{W}_{l,c,k,i,j}}$ is a 4D
tensor. As in~\eqref{eq:grad_conv2d_derivation}, an application of chain
rule yields
\begin{align*}
  &\pdv{L}{\mathcal{W}_{l,c,k,i,j}}\\
  &=\pdv{L}{\mathcal{Z}_{l,*,*,*}}\pdv{\mathcal{Z}_{l,*,*,*}}{\mathcal{W}_{l,c,k,i,j}}\\
  &=\sum_{o'=1}^{d_D+1}\sum_{m'=1}^{d_H+1}\sum_{n'=1}^{d_W+1}\pdv{L}{\mathcal{Z}_{l,o',m',n'}}\mathcal{X}_{c,k+o'-1,i+m'-1,j+n'-1}\\
  &=\pdv{L}{\mathcal{Z}_{l,*,*,*}}\star \mathcal{X}_{c,k:k+d_D,i:i+d_H,j:j+d_W}\,.
\end{align*}
Again, this implies that the gradient of 3D convolution can also be
obtained from 3D convolutions.
\begin{algorithm}[tp]  
  \DontPrintSemicolon
  \KwIn{gradient w.r.t. pre-activation $\pdv{L}{\mathcal{Z}}$
    of shape $[\tau, c_{\mathrm{out}}, d_H{+}1, d_W{+}1]$,
    input image $\mathcal{X}$ of shape $[\tau, c_{\mathrm{in}}, s_H, s_W]$}
  Construct $\mathcal{P}\gets \mathsf{im2col}(\mathcal{X}, [\kappa, \kappa])$\;
  \tcp{$\mathcal{P}$ is of shape $[\tau, (d_H{+}1)(d_W{+}1), \kappa^2c_{\mathrm{in}}]$}
  $\delta \mathcal{Z} \gets $ reshape $\pdv{L}{\mathcal{Z}}$ into
  $[\tau, c_{\mathrm{out}}, (d_H{+}1)(d_W{+}1)]$\;
  \tcc{Compute batch matrix-matrix multiplication}
  $\mathcal{G}$ = \textsf{bmm}($\delta\mathcal{Z}$, $\mathcal{P}$)\;
  $\mathcal{G}\gets$ reshape $\mathcal{G}$ into $[\tau, c_{\mathrm{out}}, c_{\mathrm{in}},\kappa, \kappa]$\;
  \Return $\mathcal{G}$\;
  \caption{Per-example gradient computation for convolutional layer}
  \label{alg:pegrad_conv_layer}
\end{algorithm}

\subsection{Recurrent Layers}
\begin{figure}[tp]
  \centering
  \begin{tikzpicture}[%
    mynode/.style={rectangle,rounded corners=2pt,draw,thick,align=center,
      ,minimum width=33pt,minimum height=20pt}
    ]%
    \node[mynode] (At-1) {$\vec{h}^{(t-1)}$};
    \node[mynode,right=3em of At-1] (At) {$\vec{h}^{(t)}$};
    \node[mynode,right=3em of At] (At+1) {$\vec{h}^{(t+1)}$};
    \draw[-latex] ($(At-1.west)-(3em,0)$) -- node[above] {$W$}(At-1);
    \draw[-latex] (At-1)-- node[above] {$W$} (At);
    \draw[-latex] (At) -- node[above] {$W$} (At+1);
    \draw[-latex] (At+1) --  ($(At+1.east)+(3em,0)$);
    \node[draw=none,below=3em of At-1] (xt-1) {$\vec{x}^{(t-1)}$};
    \node[draw=none,below=3em of At] (xt) {$\vec{x}^{(t)}$};
    \node[draw=none,below=3em of At+1] (xt+1) {$\vec{x}^{(t+1)}$};
    \draw[-latex] (xt-1) -- node[right] {$V$} (At-1);
    \draw[-latex] (xt) -- node[right] {$V$} (At);
    \draw[-latex] (xt+1) -- node[right] {$V$} (At+1);
  \end{tikzpicture}
  \caption{Recurrent neural network}
  \label{fig:recurrent_layer}
\end{figure}
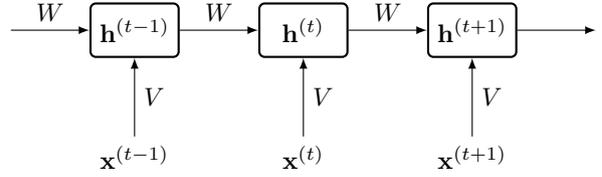
We now consider a recurrent layer with weight matrices $W \in
\R^{m\times m}$ and $V \in \R^{m\times n}$. Let $\vec{x}^{(t)} \in \R^n$ and
$\vec{h}^{(t)}\in \R^m$, for $t=1, \ldots, T$, denote the input and hidden
state vectors at time step $t$, 
respectively. As shown in Figure~\ref{fig:recurrent_layer}, the
pre-activation $\vec{z}^{(t)} \in \R^m$ at time $t$ is computed by  
\begin{align}
  \vec{z}^{(t)}
  &= W \vec{h}^{(t-1)} + V \vec{x}^{(t)} + \vec{b}\,, \label{eq:rnn_preact}
\end{align}
where $\vec{h}^{(t-1)} = \phi(\vec{z}^{(t-1)})$ and $\phi$ is an
activation function.
We first consider the gradient with respect to $W$, the weight matrix
for hidden state vector.
By the chain rule, the gradient of $L$ with respect to $W_{i,j}$ is
\begin{align}
  \pdv{L}{W_{i,j}}
  &= \sum_{t=1}^T \pdv{L}{\vec{z}^{(t)}}\pdv{\vec{z}^{(t)}}{W_{i,j}} \nonumber \\
  &=\sum_{t=1}^T\sum_{k=1}^m \pdv{L}{\vec{z}^{(t)}[k]}
    \pdv{\vec{z}^{(t)}[k]}{W_{i,j}} \\
  &= \sum_{t=1}^T\pdv{L}{\vec{z}^{(t)}[i]}\vec{h}^{(t-1)}[j] \label{eq:rnn_dldw}
  \intertext{since we have}
  \pdv{\vec{z}^{(t)}[p]}{W_{i,j}}
  &= \pdv{}{W_{i,j}}\left(\sum_{k=1}^{m} W_{p, k}\vec{h}^{(t-1)}[k] +
    \vec{b}[p]\right) \nonumber\\
  &=\begin{cases}
    \vec{h}^{(t-1)}[j] & \mbox{ if $p=i$,}\\
    0 & \mbox{ if $p\neq i$.} 
  \end{cases} \nonumber
\end{align}
From~\eqref{eq:rnn_dldw}, we see that
\begin{equation} \label{eq:recurrent_gradient}
  \pdv{L}{W}
  = \sum_{t=1}^T \pdv{L}{\vec{z}^{(t)}} \otimes \vec{h}^{(t-1)}\,.
\end{equation}
Similarly, the gradient with respect to $V$, weight matrix for input
vector, can be obtained as follows:
\begin{align*}
  \pdv{L}{V}
  &= \sum_{t=1}^T \pdv{L}{\vec{z}^{(t)}} \otimes \vec{x}^{(t)} \,\text{ and  }\,
    \pdv{L}{\vec{b}} = \sum_{t=1}^T \pdv{L}{\vec{z}^{(t)}}\,.
\end{align*}
Algorithm~\ref{alg:pegrad_recurrent_layer} describes how per-example
gradients are computed using Equation~\eqref{eq:recurrent_gradient}.
\begin{algorithm}[tp]  
  \DontPrintSemicolon
  \KwIn{list of gradient w.r.t. pre-activations
    $[\pdv{L}{\vec{z}^{(1)}}, \ldots, \pdv{L}{\vec{z}^{(T)}}]$ for $T$
    time steps,
    input sequence $[\vec{x}^{(1)}, \ldots, \vec{x}^{(T)}]$}
  $\Gamma = [\,]$\;
  \For{$t=1$ \KwTo $T$}{%
    $\mathcal{Z} \gets$ shape batch of $\pdv{L}{\vec{z}^{(t)}}$ into
    $[\tau, m, 1]$\;
    $\mathcal{X} \gets$ shape batch of $\vec{x}^{(t)}$ into $[\tau, 1, n]$\;
    $\Gamma$.\textsf{append}(\textsf{bmm}($\mathcal{Z}$, $\mathcal{X}$))\;
  }
  $\mathcal{G} \gets$ \textsf{sum}($\Gamma$)\;
  \Return $\mathcal{G}$\;
  \caption{Per-example gradient computation for recurrent layer}
  \label{alg:pegrad_recurrent_layer}
\end{algorithm}

\subsection{LSTM Layers}
The forward phase of an LSTM layer is described by the following
pre-activations 
\begin{align*}
  \begin{bmatrix}
    \vec{z}_f^{(t)} \\ \vec{z}_i^{(t)} \\ \vec{z}_g^{(t)} \\  \vec{z}_o^{(t)}\\
  \end{bmatrix}
  &=
    \begin{bmatrix}
      W^f \\ W^i \\ W^g \\ W^o 
    \end{bmatrix}
  \vec{h}^{(t-1)} +
  \begin{bmatrix}
    V^f \\ V^i \\ V^g \\ V^o 
  \end{bmatrix} \vec{x}^{(t)} +
  \begin{bmatrix}
    \vec{b}^f \\ \vec{b}^i \\ \vec{b}^g \\ \vec{b}^o\\
  \end{bmatrix}
\end{align*}
and 4 gate values
$ \vec{f}^{(t)} = \sigma(\vec{z}_f^{(t)})$, $\vec{i}^{(t)} = \sigma(\vec{z}^{(t)}_i)$,
$\vec{g}^{(t)} = \tanh(\vec{z}_g^{(t)})$,
and $\vec{o}^{(t)} = \sigma(\vec{z}_o^{(t)})$, where $\sigma(\cdot)$ is the
sigmoid function, $W_{\xi} \in \R^{m\times m}$ and $V_\xi \in
\R^{m\times n}$ for $\xi \in \{f, i, g, o\}$.
The above can be simplified by introducing matrices $W \in
\R^{4m\times m}$ and $V \in \ R^{4m \times n}$ and a bias $\vec{b} \in
\R^{4m}$ constructed by stacking weights and biases for all gates:
\[
  \vec{z}^{(t)} = W\vec{h}^{(t-1)} + V\vec{x}^{(t)} + \vec{b}\,.
\]
From the above, we see that the gradient of an LSTM layer can be
computed in the same way as in a recurrent layer.

\subsection{LayerNorm Layers}\label{subsec:layernorm}
LayerNorm \cite{layernorm} layer enables a neural network to control the distribution
of layer inputs by allowing it to control the mean and variance of
inputs across activations (rather than those across minibatch as in
batch normalization). It has two parameters $\vec{\gamma}$ and $\vec{\beta}$. In
the forward phase, the LayerNorm at layer $\ell$ computes the mean
and variance of activations from the layer $\ell-1$:
\begin{align*}
  \mu^{(\ell)}
  &= \frac{1}{\kappa} \sum_{i=1}^\kappa \vec{a}^{(\ell-1)}[i] \text{ and }
    \sigma^{(\ell)} = \frac{1}{\kappa}\sum_{i=1}^\kappa (\vec{h}^{(\ell-1)}[i] - \mu^{(\ell)})^2\,.
\end{align*}
It then normalizes the layer inputs by
\[
  \overline{\vec{h}}^{(\ell)}[i] =
  \frac{1}{\sigma^{(\ell)}}(h^{(\ell-1)}[i] - \mu^{(\ell)})\,, \quad
  i=1, \ldots, \kappa\,.
\]
Finally, the output of layer is given by
\[
  \vec{h}^{(\ell)} = \vec{\gamma} \odot \overline{\vec{h}}^{(\ell)} + \vec{\beta}\,,
\]
where $\vec{\gamma}, \vec{\beta} \in \R^\kappa$ and $\odot$ denotes the element-wise
multiplication. If we view $\vec{h}^{(\ell)}$ as the pre-activation of layer,
we have
\begin{align*}
  \pdv{L}{\vec{\gamma}}
  &= \pdv{L}{\vec{h}^{(\ell)}} \pdv{\vec{h}^{(\ell)}}{\vec{\gamma}} =
    \pdv{L}{\vec{h}^{(\ell)}} \mathrm{diag}(\overline{\vec{h}}^{(\ell)})
  = \pdv{L}{\vec{h}^{(\ell)}} \odot \overline{\vec{h}}^{(\ell)} \\
  \pdv{L}{\vec{\beta}}
  &=\pdv{L}{\vec{h}^{(\ell)}} \pdv{\vec{h}^{(\ell)}}{\vec{\beta}} = \pdv{L}{\vec{h}^{(\ell)}}
    I_k = \pdv{L}{\vec{h}^{(\ell)}}\\
  \intertext{since we have }
  \pdv{\vec{h}^{(\ell)}[p]}{\vec{\gamma}[i]}
  &=\pdv{}{\vec{\gamma}[i]}\left(\vec{\gamma}[p]\overline{\vec{h}}^{(\ell)}[p] +
    \vec{\beta}[p]\right)
  =
    \begin{cases}
      \overline{\vec{h}}^{(\ell)}[i] & \mbox{ if $p=i$,} \\
      0 & \mbox{ if $p\neq i$,}
    \end{cases}\\
  \intertext{and}
  \pdv{\vec{h}^{(\ell)}[p]}{\vec{\beta}[i]}
  &=\pdv{}{\vec{\gamma}[i]}\left(\vec{\gamma}[p]\overline{\vec{h}}^{(\ell)}[p] +
    \vec{\beta}[p]\right)=
    \begin{cases}
      1 & \mbox{ if $p=i$,}\\
      0 & \mbox{ if $p\neq i$.}
    \end{cases}
\end{align*}
As shown in Algorithm~\ref{alg:pegrad_layernorm_layer}, the
per-example gradient for LayerNorm layer over a minibatch can be
obtained by simple element-wise product of two matrices.
\begin{algorithm}[tp]  
  \DontPrintSemicolon
  \KwIn{batch of gradient w.r.t. pre-activations
    $Z=[\pdv{L}{\vec{h}_1}^\intercal, \ldots, \pdv{L}{\vec{h}_{\tau}}^\intercal]^\intercal$,
    normalized input ${H} = [\overline{\vec{h}}_1^\intercal, \ldots, \overline{\vec{h}}_\tau^\intercal]^\intercal$}
  $\mathcal{G} \gets Z \odot H$\;
  \Return $\mathcal{G}$\;
  \caption{Per-example gradient computation for LayerNorm layer}
  \label{alg:pegrad_layernorm_layer}
\end{algorithm}

\subsection{Multi-head Attention Layers}
Multi-head attention mechanism is a core component of {\scshape Transformer}
network~\cite{Vaswani2017attention,Devlin2019bert,Yang2019xlnet}, the
state-of-the-art model for neural language translation (NLT).

\begin{figure}[tp]
  \centering
  \begin{tikzpicture}[
    cbox/.style={rectangle,anchor=west,minimum height=.7cm,
      minimum width=#1,draw=black,font=\small,},
    oplus/.style={circle,draw=black,thick,fill=white,inner sep=0pt,font=\bfseries,text=black}
    ]
    \node[cbox={4.3cm}] (lnorm1) {LayerNorm2};
    \node[cbox={4.3cm},above=.3cm of lnorm1] (fc2) {Fully-connected 2};
    \node[draw=none,font=\small,above=.3cm of fc2] (out) {Output};
    \draw[thick] (lnorm1) -- (fc2);
    \draw[thick] (fc2) -- (out);
    \node[oplus,below=.2cm of lnorm1] (op1) {+};      
    \draw[thick,black] (op1) -- (lnorm1);
    % \coordinate (mlpy) at ($(lnorm1.south)-(0,1.4)$);    
    \node[cbox={4.3cm},below=.2cm of op1] (ff)  {Fully-connected 1};
    \draw[thick,black] (op1.south) -- (ff);
    \node[cbox={4.3cm},below=.3cm of ff] (lnorm2) {LayerNorm1};
    \draw[thick,black] (ff) -- (lnorm2);
    \draw[thick,black,-latex] ($(ff.south)!0.5!(lnorm2.north)$) -- ++ (3, 0) |- (op1);
    \node[oplus,below=.2cm of lnorm2] (op2) {+};
    \node[cbox={4.3cm},below=.2cm of op2] (sa) {Multihead Attention};
    \node[cbox={4.3cm},below=.5cm of sa] (pe) {Positional Encoding};
    \draw[thick,black,-latex] ($(pe)!0.5!(sa)$) -- ++ (3, 0) |- (op2);
    \node[cbox={4.3cm},below=.3cm of pe] (we) {Word Embedding};
    \node[draw=none,font=\small,below=.3cm of we] (input) {Input};
    \draw[black,thick] (lnorm2) -- (op2) -- (sa) -- (pe) -- (we) -- (input);
    \node[fit={($(sa)!0.5!(pe)$)(lnorm1)($(sa.south)+(3.3,0)$)($(sa.south)-(3.3,0)$)},
    thick,draw=blue,label={[label distance=.25cm,text
      width=2cm]50:{\footnotesize A Transformer block}}] {};
  \end{tikzpicture}
  \caption{A transformer network with a single encoder block.\label{fig:transformer}}
\end{figure}
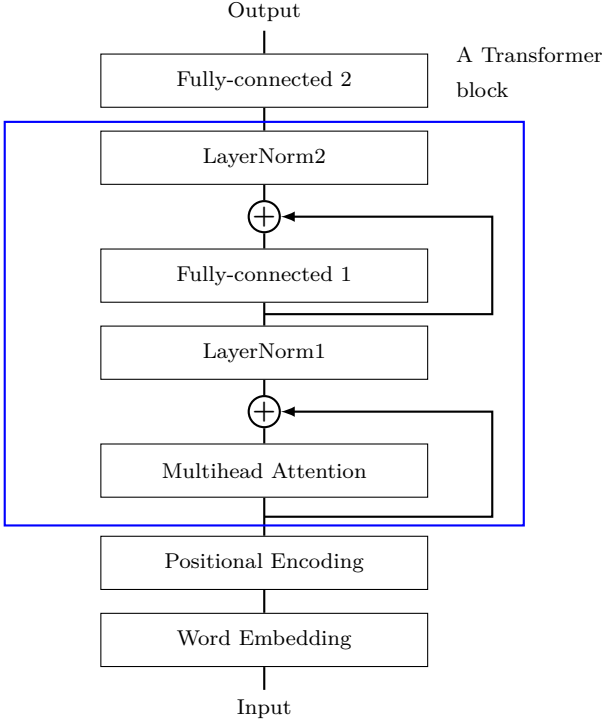

Let $X = (\vec{x}_1^\intercal, \vec{x}_2^\intercal, \ldots,
\vec{x}_s^\intercal)^\intercal$ be an input sequence of encoded vectors $\vec{x}_i
\in \R^{d_m}$, and consider a multi-head attention layer with $h$
attention heads in the $l$\th layer of a {\scshape Transformer}
network. The architecture of a transformer network with a single
encoder block is described in Figure~\ref{fig:transformer}.
The layer takes a tuple $(Q^{(\ell-1)}, K^{(\ell-1)}, V^{(\ell-1)})$
of query $Q$, key $K$, and value $V$ from the layer 
$\ell-1$ as input. Note that $(Q^{(0)}, K^{(0)}, V^{(0)}) = (X, X,
X)$. It starts by applying linear transformations on the inputs. This
is done by multiplying them with weight matrices $W^Q, W^K, W^V \in
\R^{d_m\times d_m}$: 
\begin{align*}
  Q^{(\ell)} & = Q^{(\ell-1)}(W^Q)^\intercal\,, \\
  K^{(\ell)} &= K^{(\ell-1)}(W^K)^\intercal\,, \\
  V^{(\ell)} & = V^{(\ell-1)}(W^V)^\intercal\,.
\end{align*}
The attention weights are computed by the scaled dot product between
$Q$ and $K$. The attention values are weighted sums of values $V$.
\begin{align*}
  A^{(\ell)}
  & = \frac{1}{\sqrt{d_k}}\mathrm{softmax}\left(Q^{(\ell)}(K^{(\ell)})^\intercal\right)\,,\\
  H^{(\ell)}
  &= A^{(\ell)} V^{(\ell)}\,,
\end{align*}
where $d_m = h \times d_k$.
Finally, the output of layer $Y$ is obtained by applying a linear
transformation on the attention values:
\begin{align*}
  Y^{(\ell)}
  &= H^{(\ell)}(W^O)^\intercal\,,
\end{align*}
where  $W^O\in \R^{d_m\times d_m}$. 
The gradient of $L$ with respect to $W^Q$ is
\begin{align*}
  \pdv{L}{W^Q}
  &= \pdv{L}{Q^{(\ell)}}\pdv{Q^{(\ell)}}{W^Q}\,.
\end{align*}
From
\begin{align*}
  \pdv{Q^{(\ell)}_{m,n}}{W_{i,j}}
  &=\pdv{}{W_{i,j}}\left(\sum_{k=1}^{d_m} Q_{m, k}^{(\ell-1)}W_{k,n}\right)
  =
    \begin{cases}
      Q^{(\ell)}_{m, j} & \mbox{ if $n= i$,}\\
      0 & \mbox{ if $n\neq i$,}
    \end{cases}\,,
\end{align*}
we get
\begin{align*}
\pdv{Q^{(\ell)}}{W_{i,j}}
  &=
    \begin{bmatrix}
      0 & \cdots & Q_{1, j}^{(\ell-1)} & \cdots & 0 \\
      0 & \cdots & Q_{2, j}^{(\ell-1)} & \cdots & 0 \\
      \vdots & & \vdots & & \vdots \\
      0 & \cdots & Q_{s, j}^{(\ell-1)} & \cdots & 0 \\
    \end{bmatrix}\,, \\
 \intertext{where we only have non-zero entries at the $i$\th
  column. From the above, we have}
  \pdv{L}{W_{i,j}}
  &=\sum_{k=1}^{s} \pdv{L}{Q^{(\ell)}_{k, i}} Q^{(\ell-1)}_{k,j} =
    \ip*{\pdv{L}{Q^{(\ell)}_{*,i}},\, Q^{(\ell-1)}_{*,j}}\,.
\end{align*}
In other words, the entry of $\pdv{L}{W}$ at location $(i, j)$ is
obtained by taking inner product between the $i$\th column of
$\pdv{L}{Q^{(\ell)}}$ and the $j$\th column of $Q^{(\ell-1)}$.
Combining all together, we conclude that 
\[
  \pdv{L}{W^Q} = \left(\pdv{L}{Q^{(\ell)}}\right)^\intercal Q^{(\ell-1)}\,.
\]
Similarly, we can compute the gradients with respect to other parameters:
\[
  \begin{aligned}
  \pdv{L}{W^K} &= \left(\pdv{L}{K^{(\ell)}}\right)^\intercal K^{(\ell-1)}\,, \\
  \pdv{L}{W^V} &= \left(\pdv{L}{V^{(\ell)}}\right)^\intercal V^{(\ell-1)}\,, \\
  \pdv{L}{W^O} &= \left(\pdv{L}{Y^{(\ell)}}\right)^\intercal H^{(\ell)}\,. 
\end{aligned}
\]

\subsection{Other Layer Types}

There are many types of layers that have no parameters at all. Some common examples include max-pooling layers (which divide the layer input into patches and output the maximum of each patch), softmax layers (which take a vector $\vec{x}\in \R^{d_\textrm{in}}$ and output a vector $\left[\frac{e^{\vec{x}[1]}}{\sum_i e^{\vec{x}[i]}},\cdots, \frac{e^{\vec{x}[d_{\textrm{in}}]}}{\sum_i e^{\vec{x}[i]}}\right]$). These layers do not outwardly affect our approach -- they are automatically accounted for when we ask the auto-differentiation software to compute $\pdv{\text{Loss}}{Z^{(\ell)}}$ for layers $\ell$ below them.

Similarly, skip-connections, which are used in residual blocks \cite{resnet} also do not outwardly affect our approach.

\subsection{Implementation}
We implemented the fast per-example gradient clipping technique,
described in Section~\ref{sec:algorithm}, using PyTorch. We
encapsulated the per-example gradient norm computation functionality
into python wrapper classes for PyTorch's built-in network layers,
e.g., Linear, Conv2D, RNN, and so on. This modular implementation
allows users to incorporate the gradient clipping functionality into
their existing neural network models by simply replacing their layers
with our wrapper classes. Each layer wrapper class maintains
references to two tensors: pre-activations $Z$ and input $X$ to 
the layer. After the feed-forward step, it computes $\partial L/
\partial Z$, the gradient with respect to $Z$, using \texttt{autograd}
package and combines it with
$X$ to derive per-example gradients.

%%% Local Variables:
%%% mode: latex
%%% TeX-master: "main"
%%% End:

%% file: experiments.tex
\subsection{Experimental Setup}
To evaluate the efficiency of the proposed framework, we compare the 
performance of our per-example loss reweighting algorithm to those of
two other algorithms, namely \textsf{Non-private} and \textsf{nxBP},
on different types of neural network models.
\textsf{Non-private} algorithm takes a minibatch of examples and
performs the forward and backward propagation steps only once as in
standard training process. \textsf{nxBP} is the baseline differentially private deep learning algorithm that computes per-example gradient clipping using the naive method from Section \ref{sec:clipping}: it  uses auto-differentiation to sequentially obtain the gradient for each record, clips it, and then adds the clipped gradients together. \textsf{multiLoss} is an improved version of the naive approach. As described in Section \ref{sec:clipping}, it asks the auto-differentiator to get the gradients for all examples at once (e.g., it calls \texttt{torch.autograd.grad} with first parameter equal to the vector of losses across mini-batch records) and then clips and adds them together.
Our algorithm, \textsf{ReweightGP}, performs back-propagation
twice, once for computing per-example gradient norms (as explained in Section \ref{sec:algorithm}) to determine the
weights for individual loss functions and the other for computing the
batch gradient of weighted loss function.

\textbf{We note that accuracy comparisons among the differentially private algorithms are irrelevant, as they all produce the same clipped gradients -- the only difference among them is speed.}

We have implemented our algorithm using
PyTorch~\cite{Paszke2017automatic} framework. We used a differentially 
private version of Adam optimizer, which is the same with the
non-private Adam~\cite{Kingma2014adam} except it injects Gaussian
noise with scale $\sigma$ to gradients. In our experiments, we set
the default value for the clipping threshold $C$ to be $1$ and used
the default value of $\sigma=0.05$. For all experiments, we set the
step size of Adam optimizer to 0.001, $\beta_1=0.9$, and
$\beta_2=0.999$. At each epoch, we randomly shuffle 
the dataset and partition the data into non-overlapping chunks of size
$|B|$. All the experiments were conducted on a machine with Intel Xeon
E5-2660 CPU and NVIDIA GeForce 1080 TI GPU.

\subsubsection{Models}
\label{sec:model_desc}
We tested the effectiveness of our framework on the following 5
different neural network models for classification. All models apply
softmax function to the output layers and use the cross entropy loss.
\begin{itemize}[leftmargin=*]
\item MLP (Multi-layer Perceptron): this is a simple neural network
  with two hidden layers. The first layer contains 128 and the
  second layer 256 units. We used sigmoid function as our default
  activation function.
\item CNN (Convolution Neural Network): the network consists of 2
  convolutional layers, each of which followed by a $2\times 2$ max
  pooling layer with stride of 2, and one fully connected layer with 128
  hidden units. The first convolutional layer has 20 kernels of size
  $5\times 5$ with stride 1, and the second layer 50 kernels of size
  $5\times 5$ with stride 1. We didn't use zero-paddings.
\item RNN (Recurrent Neural Network): this network was constructed by
  adding a fully connected layer on top of one vanilla recurrent layer
  with 128 hidden units. $\tanh$ was used as an activation function.
\item LSTM (Long Short-term Memory): similar to RNN, there is
  one LSTM layer with 128 hidden units followed by a fully connected
  layer for classification.
\item Transformer: the network contains a word embedding layer,
  positional encoding layer, a transformer encoder block, and a fully
  connected layer. Figure~\ref{fig:transformer} describes the architecture
  of the Transformer network used in our experiments. 
\end{itemize}

\subsubsection{Datasets and Tasks}
We used the following five publicly available datasets in our
experiments.

\begin{figure*}[ht]
  \centering
  \begin{subfigure}[b]{.76\textwidth}
    \includegraphics[width=\textwidth]{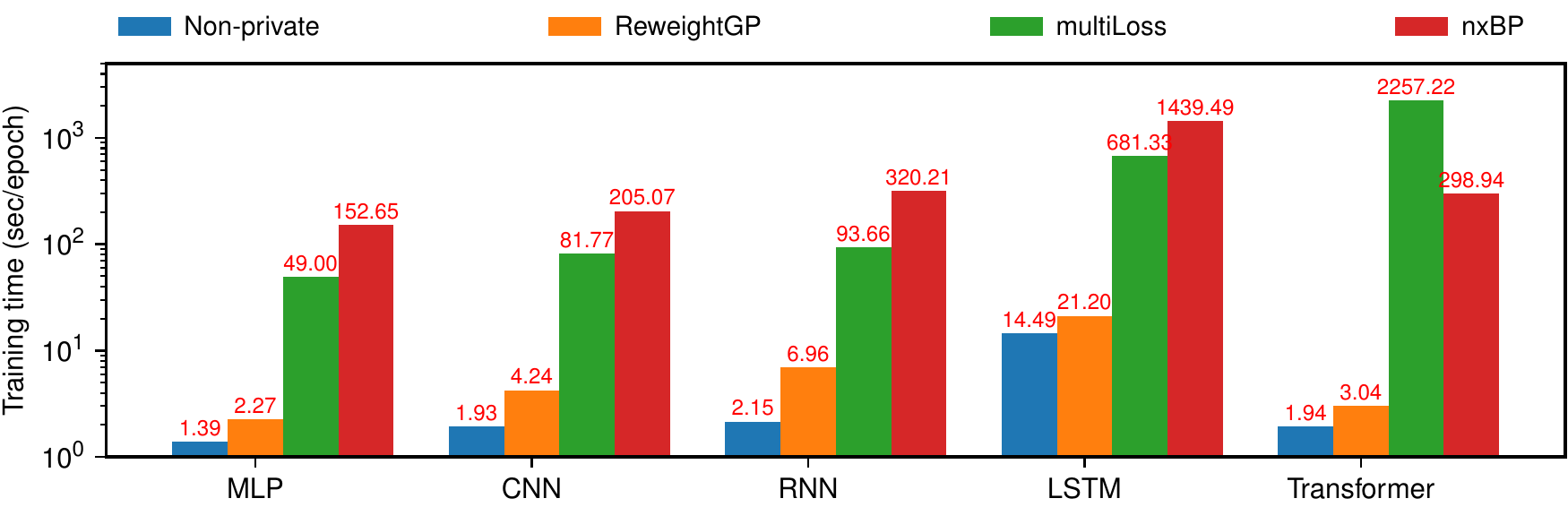}
  \end{subfigure}  
  \begin{subfigure}[b]{.38\textwidth}
    \includegraphics[width=\textwidth]{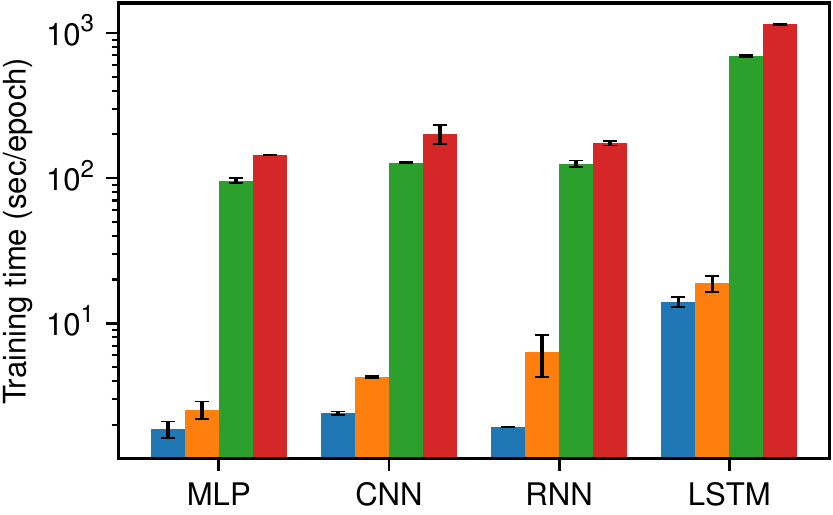}
  \end{subfigure}\quad
  \begin{subfigure}[b]{.38\textwidth}
    \includegraphics[width=\textwidth]{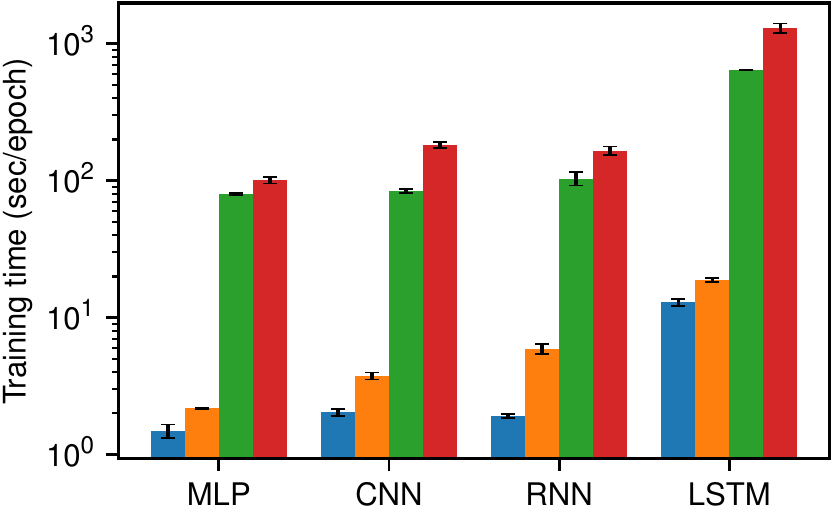}
  \end{subfigure}
  \caption{Comparison of performance by varying architectures (Top:
    MNIST, Bottom-left: FMNIST, Bottom-right: CIFAR10, Transformer is trained on IMDB)}
  \label{fig:comp_by_architecture}
\end{figure*}

\begin{enumerate}[labelindent=0pt]
\item \textbf{MNIST} is a grayscale, image dataset of hand-written
  digits, consisting of  60,000 training and 10,000 test
  examples. Each image has $28\times 28$ pixels, and there are 10
  classes (one for each digit). We trained MLP, CNN, RNN, and LSTM
  networks for classfication. 
  For RNN and LSTM, we construct a
  sequence by considering the $i$\th row of an image as an input
  vector for the time step $i$. In 
  other words, we view an image as a sequence of rows.
\item \textbf{FMNIST} (Fashion-MNIST) is a dataset of fashion article
  images designed to replace MNIST dataset. It also contains 70,000
  grayscale images of size $28\times 28$ (60,000 for training and
  10,000 for testing). 
\item \textbf{CIFAR10} is an image dataset for object
  classification. It consists of 50,000 training examples of
  $32\times 32$ RGB images. There are 10 classes, and each class has
  5,000 images.
\item \textbf{IMDB} is a movie review dataset for binary sentiment
  analysis. We trained the Transformer network on this dataset using
  50\% of examples. The other 50\% of examples were used for
  testing. For word embedding, rather than training from scratch, we
  leveraged GloVe embedding vectors of 200 dimensions, pretrained on 6
  billions of tokens.
\item \revision{\textbf{LSUN}~\cite{Yu2015LSUN} is a large-scale
    scene understanding dataset, having over 59 million RGB images of size at
  least 256$\times$ 256, and 10 different scene categories.}
\end{enumerate}

\subsection{Small Image Performance}
\revision{We first show improvements for each architecture on the smaller image datasets (MNIST, FMNIST, CIFAR10). These datasets are not appropriate for Transformer, so we use IMDB for this architecture.}
Figure~\ref{fig:comp_by_architecture} compares the performance of \revision{the}
different gradient clipping \revision{computation methods} on 5 different neural network
models in terms of training time per epoch. For this experiment, the
minibatch size $|B|$ was fixed to 32, and the models were trained  for 100 epochs. As shown in the Figure \ref{fig:comp_by_architecture}, the proposed
\textsf{RewieghtGP} algorithm significantly reduces the training time
on all 5 different architectures. Notice that values on $y$-axis are in
log scale. It is worth noting that the training of LSTM network
takes significantly longer than that for other networks
because  the per-gradient computation must access
each layer's pre-activations and input tensor. This prevents us from using
highly optimized fast implementation of LSTM such as NVIDIA's cuDNN
LSTM. For RNN, this limitation can be avoided as one can derive the
gradient of loss function with respect to pre-activations from the
gradient with respect to activations using the chain rule. 
%The bottom
%graph in Figure~\ref{fig:comp_by_architecture} shows the test
%accuracies of \textsf{ReweightGP} and \textsf{nxBP} algorithms. They
%achieve essentially the accuracy because they shares the same gradients.
%

\begin{figure*}[ht]
  \centering
  % \begin{subfigure}[b]{.32\textwidth}
  %   \includegraphics[width=\textwidth]{etime_by_batchsize_MLP_B32E50C10SIG005.pdf}
  % \end{subfigure}
  % \begin{subfigure}[b]{.32\textwidth}
  %   \includegraphics[width=\textwidth]{etime_by_batchsize_MLP_B32E50C10SIG005.pdf}
  % \end{subfigure}
  % \begin{subfigure}[b]{.32\textwidth}
  %   \includegraphics[width=\textwidth]{etime_by_batchsize_MLP_B32E50C10SIG005.pdf}
  % \end{subfigure}
  \includegraphics[width=.9\textwidth]{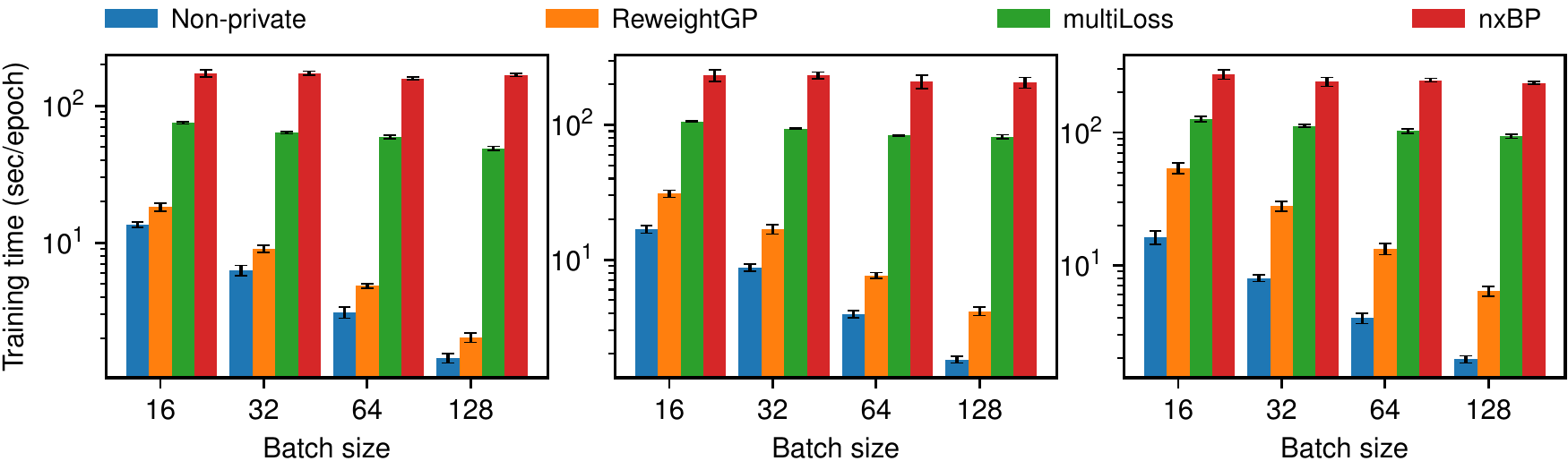}
  \caption{Execution time by varying batch size (Left: MLP, Middle:
    CNN, Right: RNN)}
  \label{fig:comp_by_batchsize}
\end{figure*}

\subsection{Impact of Different Batch Size}
Figure~\ref{fig:comp_by_batchsize} shows the impact of different batch
sizes on the per-epoch training time. For this experiment, we trained the
MLP, CNN, and RNN models described in Section~\ref{sec:model_desc} on
MNIST dataset by varying the batch size. The batch sizes used for
training are 16, 32, 64, and 128. An interesting observation is that
for \textsf{Non-private} and \textsf{ReweightGP} per-epoch training
time decreases as the batch size increases, while that 
for \textsf{nxBP} remains constant regardless of batch size. This is because
that both \textsf{Non-private} and \textsf{ReweightGP} can take
advantage of more parallelism due to the use of larger batch. On the
other hand, in \textsf{nxBP} computationally heavy error
back-propagation happens for each training example (even if an entire batch is stored in the gpu).

\subsection{Impact of Network Depth}
\label{sec:exp_network_depth}
\begin{figure*}[h]
  \centering
  \includegraphics[width=.8\textwidth]{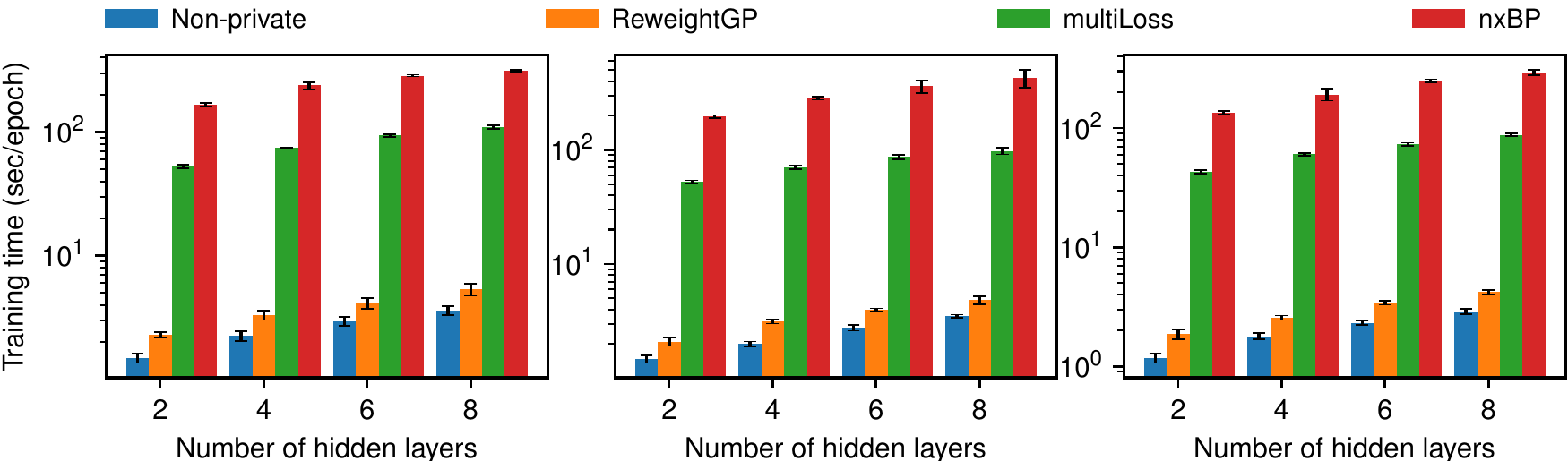}  
  % \includegraphics[width=.95\textwidth]{etimeb_by_layers_MLP_mnist_B128E30C10SIG005.pdf}
  % \begin{subfigure}[b]{.32\textwidth}
  %   \includegraphics[width=\textwidth]{etime_by_layers_MLP_mnist_B128E30C10SIG005.pdf}    
  % \end{subfigure}
  % \begin{subfigure}[b]{.32\textwidth}
  %   \includegraphics[width=\textwidth]{etime_by_layers_MLP_fmnist_B128E30C10SIG005.pdf}    
  % \end{subfigure}  
  % \begin{subfigure}[b]{.32\textwidth}
  %   \includegraphics[width=\textwidth]{etime_by_layers_MLP_cifar10_B128E30C10SIG005.pdf}    
  % \end{subfigure}
  \caption{Comparison of performance by varying number of hidden
    layers (Left: MNIST, Middle: FMNIST, Right: CIFAR10)}
  \label{fig:comp_by_depth}
\end{figure*}
\revision{Before experimenting with larger and more complex architectures,
we first provide network depth results for smaller architectures, as small networks are most commonly used
with differential privacy \cite{Abadi2016deep}.}
We trained multiple MLP models on three datasets
(MNIST, FMNIST, and CIFAR10)
by using different numbers of hiddne layers:  2, 4, 6, and 8. The batch size $|B|$
is fixed to 128. As shown in Figure~\ref{fig:comp_by_depth},
\textsf{ReweightGP} algorithm significantly outperforms the naive
\textsf{nxBP} algorithm on all three datasets. Especially on FMNIST
dataset with 2 hidden layers, the proposed algorithm showed 94x
speed-up over the naive \textsf{nxBP} algorithm.
%\revision{%
%  However, we observed that the speed-up decreases as the
%  depth increases. For example, the speed-ups for depth 2, 4, 6, and 8
%  on MNIST dataset are 72.8, 72.2, 68.8, and 58.6 respectively. This
%  is because \textsf{ReweightGP} backprops twice. However, it doesn't
%  decrease linearly because the the second
%  backprop reuses the computational graph built for the first backprop
%  and hence can be done efficiently than usual.
%}
% \begin{figure*}[t]
%   \begin{subfigure}[b]{.33\textwidth}
%     \includegraphics[width=\textwidth]{etime_by_imgsize_resnet18_lsun_B32E100C30SIG005.pdf}    
%   \end{subfigure}
%   \begin{subfigure}[b]{.33\textwidth}
%     \includegraphics[width=\textwidth]{etime_max_batch_E100C30SIG005IMG256.pdf}
%   \end{subfigure}
% \end{figure*}

\begin{figure*}[t]
  \centering
  \begin{subfigure}[b]{.31\textwidth}
    \centering
    \includegraphics[width=\textwidth]{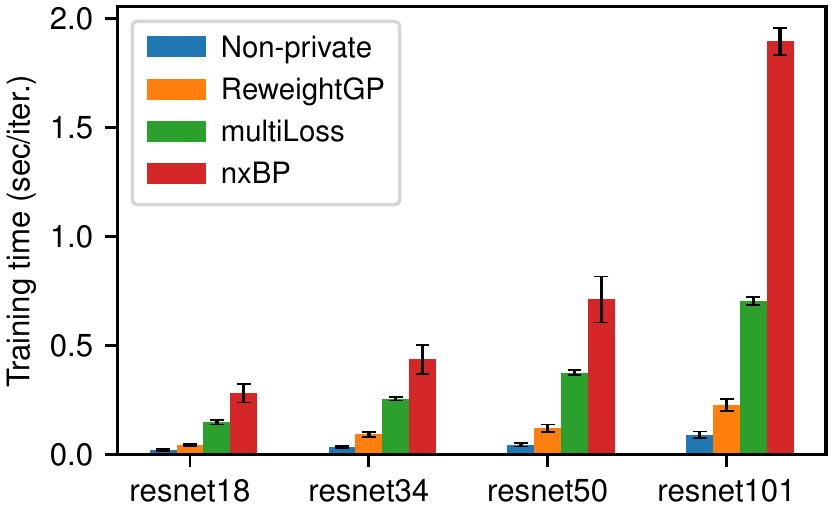}
    \caption{ResNet (Image size: 64$\times$64)}
    \label{fig:resnet_pt_64}
  \end{subfigure}
  % \begin{subfigure}[b]{.32\textwidth}
  %   \centering
  %   \includegraphics[width=\textwidth]{mem_resnet_lsun_B20E100C30SIG005IMG64.pdf}
  %   \caption{GPU memory usage (image size=64$\times$64)}
  % \end{subfigure}
  \begin{subfigure}[b]{.31\textwidth}
    \includegraphics[width=\textwidth]{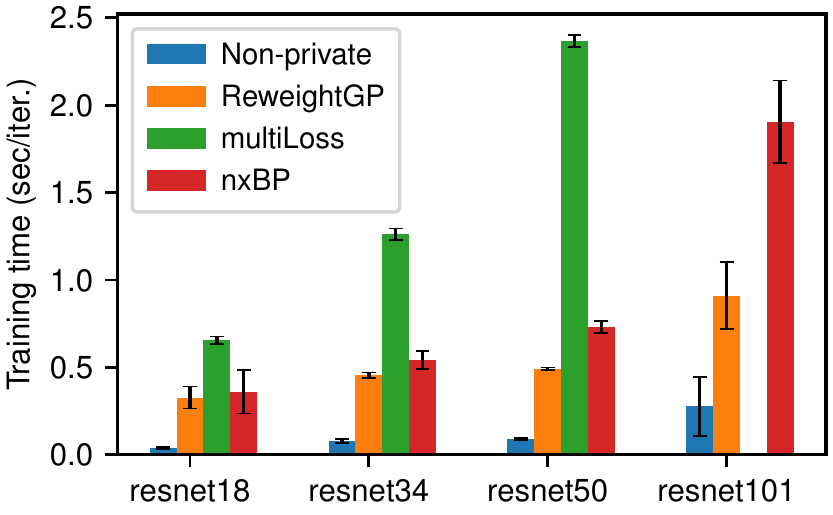}
    \caption{ResNet (Image size: 256$\times$256)}
    \label{fig:resnet_pt_256}
  \end{subfigure}
  % \begin{subfigure}[b]{.32\textwidth}
  %   \includegraphics[width=\textwidth]{mem_resnet_lsun_B20E100C30SIG005IMG256.pdf}
  %   \caption{GPU meory usage (image size=256$\times$256)}
  % \end{subfigure}
  \begin{subfigure}[b]{.31\textwidth}
    \includegraphics[width=\textwidth]{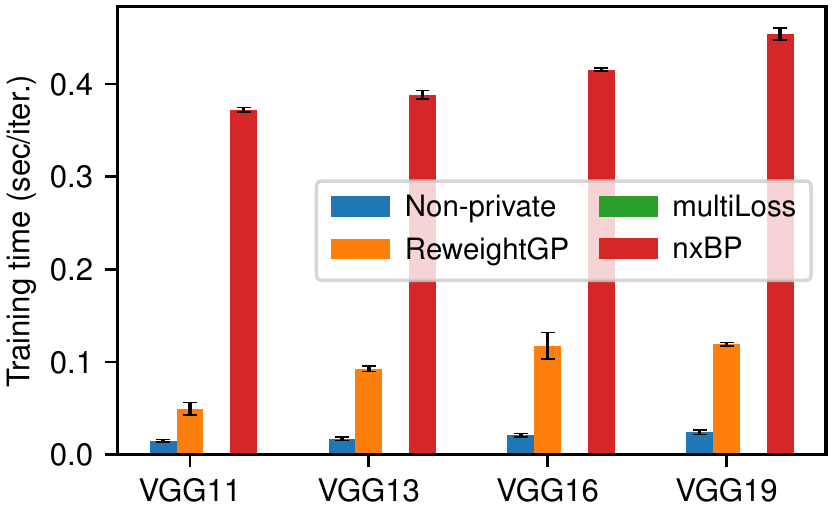}
    \caption{VGG (Image size: 64$\times$ 64)}
  \end{subfigure}
  \caption{Performance evaluation on ResNet and VGG networks. Bar for multiLoss is missing when it runs out of memory.}
  \label{fig:perf_resnet}
\end{figure*}

% \begin{figure}
%   \centering  
%   \begin{subfigure}[b]{.33\textwidth}
%     \includegraphics[width=\textwidth]{mem_vgg_lsun_B16E100C30SIG005IMG64.pdf}
%   \end{subfigure}
%   \caption{Performance evaluation on VGG networks (Top: processing
%     time, Bottom: GPU memory usage, batch size=16, image size=64$\times$ 64)}
% \end{figure}

\subsection{\textcolor{blue}{ResNet and VGG Networks}}
\revision{%
We now evaluate the performance on deeper architectures with millions of parameters: ResNet~\cite{resnet} and VGG
networks~\cite{simonyan15vgg}. For this evaluation, we froze the batchnorm parameters at values taken from pre-trained models (since batch-norm parameters do not have per-example gradients).}\footnote{\revision{ In practice, other types of normalizations could also be used, such as LayerNorm \cite{layernorm} (Section \ref{subsec:layernorm}), group norm \cite{groupnorm}, and instance norm \cite{instancenorm}}.}
%ResNet contains several batch normalization
%(BatchNorm) layers. 
%As the gradient clipping technique is not
%immediately applicable to BatchNorm layers, we exclude them from
%training. The weights for BatchNorm layers are copied from the networks
%pretrained on ImageNet dataset and frozen during the training. 
\revision{%
Due to
the large memory space requirement, mini-batches of size 20 are used for
this experiment. Results on the LSUN dataset are shown in
Figure~\ref{fig:perf_resnet}. \textsf{mutiLoss} had out-of-memory errors on VGG networks and resnet101 for large images.
% Comparing to other algorithms,
%\textsf{multiloss} algorithm requires larger memory spaces. It fails
%to run on resnet101 with the input image size is $256\times
%256$ and all VGG networks with input size 64.
%out of memory error. It also fails to run on VGG networks due to high
%memory requirement.  
We still see that \textsf{ReweightGP} consistently outperforms other gradient clipping algorithms (\textsf{nxBP}, \textsf{multiLoss}). 
The improvement is significant for images of (rescaled) size 64x64 and diminishes for size 256x256.
%outperforms other
%baselines on all architectures, its speedup over \textsf{nxBP}
%decreases on high resolution images and deeper networks.
}

\subsection{\revision{Image Size}}
\label{sec:exp_limitation}
\begin{figure}[h]
  \centering
  \includegraphics[width=.35\textwidth]{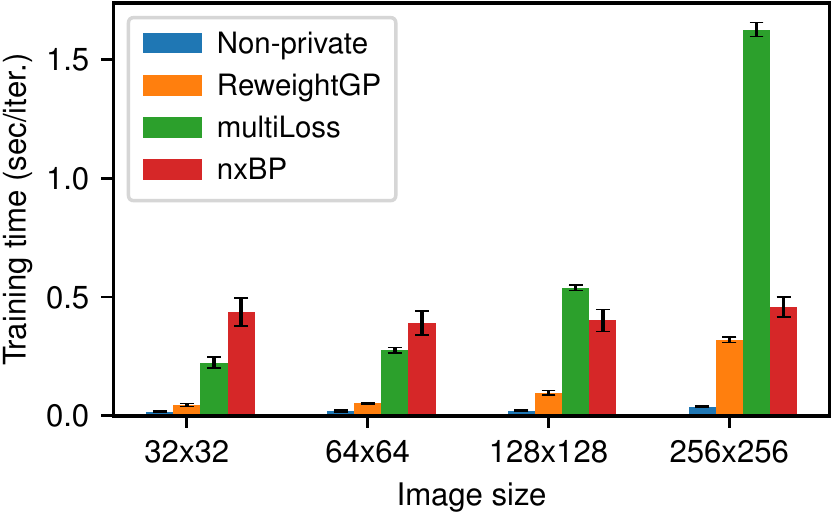}
  \caption{Processing time by image resolution}
  \label{fig:resnet_by_imgsize}
\end{figure}

\revision{
Noting that image size played a key role in reducing the speedup, we investigate this further in
  Figure~\ref{fig:resnet_by_imgsize} using ResNet 18 with batch size 32 and image sizes ranging from
  32x32 to 256x256. This causes quadratic growth in the width of the network (multiplying each dimension by c results in $c^2$ as many pixels) and we see that the advantage over the naive method decreases due to the extra computation per layer that \textsf{ReweightGP} uses.
%  
%  
%  describes how the training time
%changes as the resolution of input image increases. Both
%Figure~\ref{fig:perf_resnet} and~~\ref{fig:resnet_by_imgsize}
%suggest that the size of input (image resolution) has more impact on
%the performance than the depth of network. 
%  This is because that
%convolutional networks on high resolution input images perform linear
%algebra operations on large tensors. However, the efficiency of 
%underlying cuBLAS libarary doesn't scale linearly w.r.t. its input
%size.
}
\subsection{\revision{Memory}}
\revision{%
  Due to caching, it is difficult to obtain an accurate estimate of GPU memory requirements.
  As an alternative, we consider the largest batch size a method can support before running out of memory.
  For this experiment, we used ResNet 101 with 256x256 input images and varied the batch sizes. The non-private method
  first failed at batch size 48, \textsf{ReweightGP} at 36, and \textsf{multiLoss} at 18. \textsf{nxBP} operates on one example at a time (even when an entire batch is stored in the GPU). Thus we estimate the GPU memory overhead of \textsf{ReweightGP} compared to nonprivate to be up to $(48-36)/48\approx$ 25\% for large images. At the lower end, \textsf{ReweightGP} with ResNet 18 with 32x32 images ran with batch size of 500 without any problems. Note \textsf{nxBP}  under-utilizes GPU memory and parallelism (backpropagating through one example at a time). Thus, in practice, the memory overhead is manageable (i.e., allows for relatively large batch sizes) and buys  us significant improvements in running time (taking better advantage of GPU parallelism). 
  %
 % To compare the memory requirement of each algorithm, in
 % Figure~\ref{fig:reset101_img256}, we train resnet101 model on LSUN
 % dataset with varying batch sizes. Starting from 10,
 % we gradually increases the batch size until each algorithm fails to
 % run. Note that \textsf{nxBP} is excluded from the figure as it
 % always process one example at a time (i.e., mini-batch of size 1).
 % As shown, the proposed \textsf{ReweightGP} algorithm requires
 % more memory space than \textsf{Non-private} algorithm. This memory
 % requirement limits the size of mini-batches used for training
 % private models. 
}

%The processing time as batches are varied (up to the failure point) are shown in Figure ~\ref{fig:reset101_img256}.
%\begin{figure}[h]
%  \centering
%  \includegraphics[width=.4\textwidth]{etime_max_batch_E100C30SIG005IMG256.pdf}
%  \caption{Processing time by varying batch size}
%  \label{fig:reset101_img256}
%\end{figure}
  
\subsection{\revision{Limitations}}
\revision{Overall, the experiments have shown that our proposed \textsf{ReweightGP} method outperforms the other methods  \textsf{nxBP} and \textsf{MultiLoss} (which is often unreliable). \textsf{ReweightGP} requires more memory and computation per layer than \textsf{nxBP}. As a result, its advantage starts to decline with increased image sizes as this causes a quadratic scaling in the width of the network and consequently in the computations of \textsf{ReweightGP}. For very high resolution images, it may be preferable to use \textsf{nxBP}.}

\revision{Second, some highly optimized versions of LSTM, such as the ones that use the CuDNN LSTM routines do not expose the internal gate values, so that we cannot obtain the appropriate gradients. However, less optimized versions of LSTM can be implemented in PyTorch/TensorFlow and benefit from our approach.}
%%% Local Variables:
%%% mode: latex
%%% TeX-master: "main"
%%% End:

%% file: conclusion.tex
\revision{We presented a general framework for fast per-example gradient
clipping which can be used to improve training speed under differential privacy. 
Prior work underutilized GPU parallelism, leading to slow training times.
Our empirical
evaluation showed a significant reduction in training time of differentially private models.}

\revision{Per-example gradient clipping is not compatible with Batch Norm \cite{batchnorm}, but other
layer normalization methods can be used instead \cite{layernorm,groupnorm,instancenorm}.}
%%% Local Variables:
%%% mode: latex
%%% TeX-master: "main"
%%% End:

%% file: main.bbl
\begin{thebibliography}{10}

\bibitem{tensorflow2015-whitepaper}
M.~Abadi, A.~Agarwal, P.~Barham, E.~Brevdo, Z.~Chen, C.~Citro, G.~S. Corrado,
  A.~Davis, J.~Dean, M.~Devin, S.~Ghemawat, I.~Goodfellow, A.~Harp, G.~Irving,
  M.~Isard, Y.~Jia, R.~Jozefowicz, L.~Kaiser, M.~Kudlur, J.~Levenberg,
  D.~Man\'{e}, R.~Monga, S.~Moore, D.~Murray, C.~Olah, M.~Schuster, J.~Shlens,
  B.~Steiner, I.~Sutskever, K.~Talwar, P.~Tucker, V.~Vanhoucke, V.~Vasudevan,
  F.~Vi\'{e}gas, O.~Vinyals, P.~Warden, M.~Wattenberg, M.~Wicke, Y.~Yu, and
  X.~Zheng.
\newblock {TensorFlow}: Large-scale machine learning on heterogeneous systems,
  2015.
\newblock Software available from tensorflow.org.

\bibitem{Abadi2016deep}
M.~Abadi, A.~Chu, I.~Goodfellow, H.~B. McMahan, I.~Mironov, K.~Talwar, and
  L.~Zhang.
\newblock Deep learning with differential privacy.
\newblock In {\em Proceedings of the 2016 ACM SIGSAC Conference on Computer and
  Communications Security}, pages 308--318. ACM, 2016.

\bibitem{Abay2018:PPS}
N.~C. Abay, Y.~Zhou, M.~Kantarcioglu, B.~M. Thuraisingham, and L.~Sweeney.
\newblock Privacy preserving synthetic data release using deep learning.
\newblock In {\em Machine Learning and Knowledge Discovery in Databases -
  European Conference, {ECML} {PKDD} 2018, Dublin, Ireland, September 10-14,
  2018, Proceedings, Part {I}}, pages 510--526, 2018.

\bibitem{AcsGAN}
G.~Acs, L.~Melis, C.~Castelluccia, and E.~D. Cristofaro.
\newblock Differentially private mixture of generative neural networks.
\newblock In {\em ICDM}, 2017.

\bibitem{layernorm}
L.~J. Ba, J.~R. Kiros, and G.~E. Hinton.
\newblock Layer normalization.
\newblock {\em CoRR}, abs/1607.06450, 2016.

\bibitem{shmatfairness}
E.~Bagdasaryan and V.~Shmatikov.
\newblock Differential privacy has disparate impact on model accuracy.
\newblock {\em CoRR}, abs/1905.12101, 2019.

\bibitem{Bassily2014PERM}
R.~Bassily, A.~Smith, and A.~Thakurta.
\newblock Private empirical risk minimization: Efficient algorithms and tight
  error bounds.
\newblock In {\em Proceedings of the 2014 IEEE 55th Annual Symposium on
  Foundations of Computer Science}, FOCS '14, pages 464--473, Washington, DC,
  USA, 2014. IEEE Computer Society.

\bibitem{BeaulieuJones159756}
B.~K. Beaulieu-Jones, Z.~S. Wu, C.~Williams, R.~Lee, S.~P. Bhavnani, J.~B.
  Byrd, and C.~S. Greene.
\newblock Privacy-preserving generative deep neural networks support clinical
  data sharing.
\newblock {\em bioRxiv}, 2018.

\bibitem{Bottou2010large}
L.~Bottou.
\newblock Large-scale machine learning with stochastic gradient descent.
\newblock In {\em Proceedings of COMPSTAT'2010}, pages 177--186. Springer,
  2010.

\bibitem{Chaudhuri2011Objpert}
K.~Chaudhuri, C.~Monteleoni, and A.~D. Sarwate.
\newblock Differentially private empirical risk minimization.
\newblock {\em Journal of Machine Learning Research}, 12(Mar):1069--1109, 2011.

\bibitem{Chellapilla2006high}
K.~Chellapilla, S.~Puri, and P.~Simard.
\newblock High performance convolutional neural networks for document
  processing.
\newblock In {\em Tenth International Workshop on Frontiers in Handwriting
  Recognition}. Suvisoft, 2006.

\bibitem{Chen2019renyi}
C.~Chen, J.~Lee, and D.~Kifer.
\newblock Renyi differentially private erm for smooth objectives.
\newblock In {\em The 22nd International Conference on Artificial Intelligence
  and Statistics}, pages 2037--2046, 2019.

\bibitem{ChenDPGAN}
Q.~Chen, C.~Xiang, M.~Xue, B.~Li, N.~Borisov, D.~Kaafar, and H.~Zhu.
\newblock Differentially private data generative models.
\newblock \url{https://arxiv.org/pdf/1812.02274.pdf}, 2018.

\bibitem{Devlin2019bert}
J.~Devlin, M.-W. Chang, K.~Lee, and K.~Toutanova.
\newblock Bert: Pre-training of deep bidirectional transformers for language
  understanding.
\newblock In {\em Proceedings of the 2019 Conference of the North American
  Chapter of the Association for Computational Linguistics: Human Language
  Technologies, Volume 1 (Long and Short Papers)}, pages 4171--4186, 2019.

\bibitem{Dwork2006our}
C.~Dwork, K.~Kenthapadi, F.~McSherry, I.~Mironov, and M.~Naor.
\newblock Our data, ourselves: Privacy via distributed noise generation.
\newblock In {\em Annual International Conference on the Theory and
  Applications of Cryptographic Techniques}, pages 486--503. Springer, 2006.

\bibitem{Dwork2006calibrating}
C.~Dwork, F.~McSherry, K.~Nissim, and A.~Smith.
\newblock Calibrating noise to sensitivity in private data analysis.
\newblock In {\em Theory of Cryptography Conference}, pages 265--284. Springer,
  2006.

\bibitem{Goodfellow2015efficient}
I.~Goodfellow.
\newblock Efficient per-example gradient computations.
\newblock {\em arXiv preprint arXiv:1510.01799}, 2015.

\bibitem{GoodBengCour16}
I.~J. Goodfellow, Y.~Bengio, and A.~Courville.
\newblock {\em Deep Learning}.
\newblock MIT Press, Cambridge, MA, USA, 2016.
\newblock \url{http://www.deeplearningbook.org}.

\bibitem{resnet}
K.~He, X.~Zhang, S.~Ren, and J.~Sun.
\newblock Deep residual learning for image recognition.
\newblock In {\em CVPR}, 2016.

\bibitem{batchnorm}
S.~Ioffe and C.~Szegedy.
\newblock Batch normalization: Accelerating deep network training by reducing
  internal covariate shift.
\newblock {\em CoRR}, abs/1502.03167, 2015.

\bibitem{Iyengar2019amp}
R.~Iyengar, J.~P. Near, D.~Song, O.~Thakkar, A.~Thakurta, and L.~Wang.
\newblock Towards practical differentially private convex optimization.
\newblock In {\em Towards Practical Differentially Private Convex
  Optimization}, page~0. IEEE.

\bibitem{Jia2014learning}
Y.~Jia.
\newblock {\em Learning semantic image representations at a large scale}.
\newblock PhD thesis, UC Berkeley, 2014.

\bibitem{Jordon2019pategan}
J.~Jordon, J.~Yoon, and M.~van~der Schaar.
\newblock Pate-gan: Generating synthetic data with differential privacy
  guarantees.
\newblock In {\em ICLR}, 2019.

\bibitem{Kifer2012erm}
D.~Kifer, A.~Smith, and A.~Thakurta.
\newblock Private convex empirical risk minimization and high-dimensional
  regression.
\newblock In {\em Conference on Learning Theory}, pages 25--1, 2012.

\bibitem{Kingma2014adam}
D.~Kingma and J.~Ba.
\newblock Adam: A method for stochastic optimization.
\newblock {\em International Conference on Learning Representations}, 12 2014.

\bibitem{Lee2018cdp}
J.~Lee and D.~Kifer.
\newblock Concentrated differentially private gradient descent with adaptive
  per-iteration privacy budget.
\newblock In {\em Proceedings of the 24th ACM SIGKDD International Conference
  on Knowledge Discovery \& Data Mining}, 2018.

\bibitem{Mcmahan2018general}
H.~B. McMahan, G.~Andrew, U.~Erlingsson, S.~Chien, I.~Mironov, N.~Papernot, and
  P.~Kairouz.
\newblock A general approach to adding differential privacy to iterative
  training procedures.
\newblock {\em arXiv preprint arXiv:1812.06210}, 2018.

\bibitem{Brendan2018learning}
H.~B. McMahan, D.~Ramage, K.~Talwar, and L.~Zhang.
\newblock Learning differentially private recurrent language models.
\newblock In {\em International Conference on Learning Representations}, 2018.

\bibitem{Mironov2017renyi}
I.~Mironov.
\newblock Renyi differential privacy.
\newblock In {\em Computer Security Foundations Symposium (CSF), 2017 IEEE
  30th}, pages 263--275. IEEE, 2017.

\bibitem{pate}
N.~Papernot, M.~Abadi, Úlfar Erlingsson, I.~Goodfellow, and K.~Talwar.
\newblock Semi-supervised knowledge transfer for deep learning from private
  training data.
\newblock In {\em Proceedings of the International Conference on Learning
  Representations}, 2017.

\bibitem{tensorflowprivacy}
N.~Papernot, S.~Chien, C.~C. Choo, G.~M. Andrew, and I.~Mironov.
\newblock {TensorFlow Privacy}.

\bibitem{Papernot2018pate}
N.~Papernot, S.~Song, I.~Mironov, A.~Raghunathan, K.~Talwar, and Úlfar
  Erlingsson.
\newblock Scalable private learning with pate.
\newblock In {\em International Conference on Learning Representations (ICLR)},
  2018.

\bibitem{Paszke2017automatic}
A.~Paszke, S.~Gross, S.~Chintala, G.~Chanan, E.~Yang, Z.~DeVito, Z.~Lin,
  A.~Desmaison, L.~Antiga, and A.~Lerer.
\newblock Automatic differentiation in {PyTorch}.
\newblock In {\em NeurIPS Autodiff Workshop}, 2017.

\bibitem{PhanAAAI2016}
N.~Phan, Y.~Wang, X.~Wu, and D.~Dou.
\newblock Differential privacy preservation for deep auto-encoders: an
  application of human behavior prediction.
\newblock In {\em AAAI}, 2016.

\bibitem{KNG}
M.~Reimherr and J.~Awan.
\newblock {KNG:} the k-norm gradient mechanism.
\newblock In {\em NeurIPS}, 2019.

\bibitem{Robbins1951stochastic}
H.~Robbins and S.~Monro.
\newblock A stochastic approximation method.
\newblock {\em The annals of mathematical statistics}, pages 400--407, 1951.

\bibitem{Rochette2019EfficientPG}
G.~Rochette, A.~Manoel, and E.~W. Tramel.
\newblock Efficient per-example gradient computations in convolutional neural
  networks.
\newblock {\em ArXiv}, abs/1912.06015, 2019.

\bibitem{RuderOpt}
S.~Ruder.
\newblock An overview of gradient descent optimization algorithms.
\newblock {\em CoRR}, abs/1609.04747, 2016.

\bibitem{RezaDeep}
R.~Shokri and V.~Shmatikov.
\newblock Privacy-preserving deep learning.
\newblock In {\em Proceedings of the 22Nd ACM SIGSAC Conference on Computer and
  Communications Security}, 2015.

\bibitem{membershipinference}
R.~{Shokri}, M.~{Stronati}, C.~{Song}, and V.~{Shmatikov}.
\newblock Membership inference attacks against machine learning models.
\newblock In {\em IEEE Symposium on Security and Privacy (SP)}, 2017.

\bibitem{simonyan15vgg}
K.~Simonyan and A.~Zisserman.
\newblock Very deep convolutional networks for large-scale image recognition.
\newblock In {\em International Conference on Learning Representations}, 2015.

\bibitem{thakkarclip}
O.~Thakkar, G.~Andrew, and H.~B. McMahan.
\newblock Differentially private learning with adaptive clipping.
\newblock {\em CoRR}, abs/1905.03871, 2019.

\bibitem{instancenorm}
D.~Ulyanov, A.~Vedaldi, and V.~S. Lempitsky.
\newblock Instance normalization: The missing ingredient for fast stylization.
\newblock {\em CoRR}, abs/1607.08022, 2016.

\bibitem{Vaswani2017attention}
A.~Vaswani, N.~Shazeer, N.~Parmar, J.~Uszkoreit, L.~Jones, A.~N. Gomez,
  {\L}.~Kaiser, and I.~Polosukhin.
\newblock Attention is all you need.
\newblock In {\em Advances in neural information processing systems}, pages
  5998--6008, 2017.

\bibitem{Wang2017dpsvrg}
D.~Wang, M.~Ye, and J.~Xu.
\newblock Differentially private empirical risk minimization revisited: Faster
  and more general.
\newblock In {\em Advances in Neural Information Processing Systems 30}, pages
  2719--2728. Curran Associates, Inc., 2017.

\bibitem{groupnorm}
Y.~Wu and K.~He.
\newblock Group normalization.
\newblock In {\em ECCV}, 2018.

\bibitem{Xie:DPGAN}
L.~Xie, K.~Lin, S.~Wang, F.~Wang, and J.~Zhou.
\newblock Differentially private generative adversarial network, 2018.

\bibitem{Yang2019xlnet}
Z.~Yang, Z.~Dai, Y.~Yang, J.~Carbonell, R.~Salakhutdinov, and Q.~V. Le.
\newblock Xlnet: Generalized autoregressive pretraining for language
  understanding.
\newblock {\em arXiv preprint arXiv:1906.08237}, 2019.

\bibitem{Yu2015LSUN}
F.~Yu, Y.~Zhang, S.~Song, A.~Seff, and J.~Xiao.
\newblock Lsun: Construction of a large-scale image dataset using deep learning
  with humans in the loop.
\newblock {\em ArXiv}, abs/1506.03365, 2015.

\bibitem{Yu2019modelpub}
L.~Yu, L.~Liu, C.~Pu, M.~E. Gursoy, and S.~Truex.
\newblock Differentially private model publishing for deep learning.
\newblock {\em 2019 IEEE Symposium on Security and Privacy (SP)}, pages
  332--349, 2019.

\bibitem{Zhang2017RR}
J.~Zhang, K.~Zheng, W.~Mou, and L.~Wang.
\newblock Efficient private erm for smooth objectives.
\newblock In {\em Proceedings of the 26th International Joint Conference on
  Artificial Intelligence}, pages 3922--3928. AAAI Press, 2017.

\end{thebibliography}
